\documentclass[preprint,12pt]{elsarticle}

\usepackage{amssymb}
\usepackage{amsmath}

\usepackage{hyperref}
\usepackage{cleveref}
\usepackage{multicol}
\usepackage{multirow}
\usepackage{booktabs}
\usepackage{subcaption}
\usepackage{float}
\usepackage{makecell}
\usepackage{subcaption}

\journal{Medical Image Analysis}

\begin{document}

\begin{frontmatter}



\title{SimMIL: A Universal Weakly Supervised Pre-Training Framework for Multi-Instance Learning in Whole Slide Pathology Images}


\author[1]{Yicheng Song} 

\author[2]{Tiancheng Lin}

\author[1]{Die Peng} 

\author[2]{Su Yang}

\author[1]{Yi Xu \corref{cor1}} 
\ead{xuyi@sjtu.edu.cn}

\affiliation[1]{organization={Shanghai Jiao Tong University},
            city={Shanghai},
            country={China}}

\affiliation[2]{organization={Fudan University},
            city={Shanghai},
            country={China}}
\cortext[cor1]{Corresponding author}

\begin{abstract}
Various multi-instance learning (MIL) based approaches have been developed and successfully applied to whole-slide pathological images (WSI). 
Existing MIL methods emphasize the importance of feature aggregators, but largely neglect the instance-level representation learning. 
They assume that the availability of a pre-trained feature extractor can be directly utilized or fine-tuned, which is not always the case.
This paper proposes to pre-train feature extractor for MIL via a weakly-supervised scheme, i.e., propagating the weak bag-level labels to the corresponding instances for supervised learning. 
To learn effective features for MIL, we further delve into several key components, including strong data augmentation, a non-linear prediction head and the robust loss function.
We conduct experiments on common large-scale WSI datasets and find it achieves better performance than other pre-training schemes (e.g., ImageNet pre-training and self-supervised learning) in different downstream tasks.
We further show the compatibility and scalability of the proposed scheme by deploying it in fine-tuning the pathological-specific models and pre-training on merged multiple datasets.
To our knowledge, this is the first work focusing on the representation learning for MIL.

\end{abstract}


\begin{highlights}
\item We propose to incorporate task-specific domain knowledge by designing a pretext
task based on the MIL assumption.
\item We propose SimMIL, a simple framework for representation learning on MIL.
\item We conduct preliminary experiments, indicating the necessity of 
a pre-training scheme for MIL.
\item We validate SimMIL's efficacy on multiple downstream tasks.
\item We further explore the compatibility and scalability of SimMIL.

\end{highlights}


\end{frontmatter}



\section{Introduction}
\label{sec:intro}

In 2017, the world's first whole-slide scanner (IntelliSite) was approved by the Food and Drug Administration~\cite{IntelliSite}, marking a major inflection point—computational pathology (CPath) is set to revolutionize cancer diagnosis and treatment. Recent advances in CPath have shown great promise in both basic tasks~\cite{grading,plip} and advanced tasks~\cite{mutation,MSI} for analyzing whole-slide pathological images (WSIs). Among these advances, multi-instance learning (MIL), a typical annotation-efficient learning paradigm, has been intensively studied~\cite{copathsurvey}. By taking WSIs and their corresponding patches as bags and instances, MIL addresses the WSI-specific challenges of extremely high resolution and weak annotations.

\begin{figure}[h!]
    \centering
    \begin{subfigure}{0.8\linewidth}
        \includegraphics[width=0.95\linewidth]{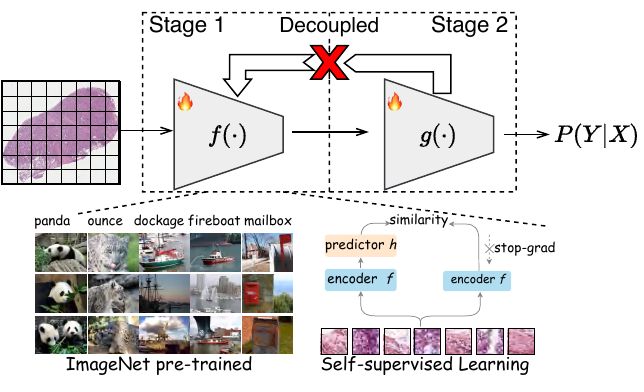}
        \caption{Decoupled}
        \label{fig:decoupled}
    \end{subfigure}
    \begin{subfigure}{0.8\linewidth}
	\includegraphics[width=0.95\linewidth]{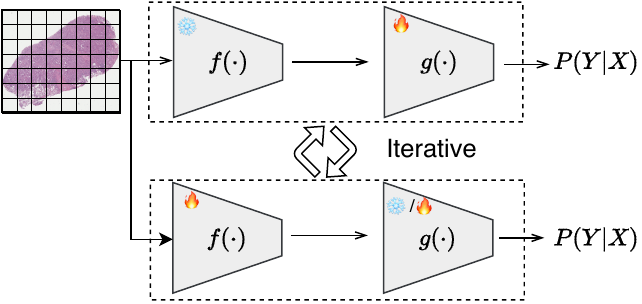}
    \caption{Iterative}
	\label{fig:iterative}
	\end{subfigure}
    \begin{subfigure}{0.8\linewidth}
        \includegraphics[width=0.95\linewidth]{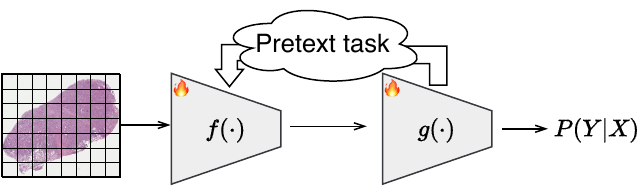}
        \caption{Pretext task-based}
        \label{fig:WSL}
    \end{subfigure}
\caption{Typical training schemes.}
\end{figure}

Most existing two-stage MIL methods focus more on designing feature aggregators while giving less attention to feature extractors~\cite{abmil,nic,dsmil,clam,transmil, dtfdmil}. This overlooks the fact that representation quality is fundamental to various downstream tasks in both natural images~\cite{moco,SimCLR} and pathological images~\cite{,uni}.
In prevalent MIL approaches, the feature extractor and feature aggregator are usually trained in a decoupled manner (see~\cref{fig:decoupled}), meaning the training of the second stage no longer affects the first stage, which can result in suboptimal solutions~\cite{cluster2conquer}.
Two types of feature extractors are commonly employed:
1) Directly applying off-the-shelf feature extractors (e.g., ImageNet pre-trained ones)~\cite{transmil,dtfdmil,yao2020whole}, which faces the issue of domain gap;
2) Pre-training the feature extractors via self-supervised learning on WSIs~\cite{dsmil,graphmil}, which introduces modality-aware domain knowledge but remains task-agnostic.
More recently, a set of methods has proposed to train the feature extractor and feature aggregator in an iterative manner, with the aim of obtaining task-specific representations.(see
~\cref{fig:iterative}). 
Given a pre-trained feature extractor, specific mechanisms are devised to perform instance selection~\cite{taskspecificFT,selfpaceIS} or pseudo-label generation~\cite{iterativePL} based on the frozen feature extractor, and these selected instances are subsequently used for fine-tuning the feature extractor.
However, this creates a chicken-and-egg dilemma—high-quality instance selection and pseudo-labels cannot be achieved without appropriate initialization of the feature extractor. Therefore, it is desirable to develop a more effective scheme for training the MIL feature extractor for WSI analysis.

In this paper, we aim to improve the representation quality of MIL as the first work focusing on the pre-training scheme for MIL tasks.
As shown in~\cref{fig:WSL}, we propose to incorporate task-specific domain knowledge by designing a pretext task, a weakly supervised pre-training scheme based on the standard MIL assumption.

Given the bag-level labels, we propagate the labels to the corresponding instances and conduct supervised learning for benign-malignant classification and cancer subtyping tasks, also known as SimpleMIL~\cite{SimpleMIL}~\footnote{A widely-used baseline for instance-level MIL.} and use supervised contrastive learning for the survival prediction task, following~\cite{panther}.

Building on this scheme, we propose \textit{SimMIL}, a simple framework for representation learning on MIL, which includes several key components, such as strong data augmentation, a non-linear prediction head, and the loss function.
\textit{SimMIL} establishes a bridge between the two counterparts of instance-level and bag-level MIL~\cite{benchmarkingmil}, offering the insight that instance-level MIL can function as a strong feature extractor for bag-level MIL methods.

We conduct preliminary experiments on the NCTCRC~\cite{NCTCRC}, and the results demonstrate that good linear probing and fine-tuning do not ensure strong MIL performance, indicating the necessity of a pre-training scheme for MIL.

We validate SimMIL's efficacy on downstream tasks, including benign-malignant classification, cancer subtyping, and survival prediction. For the first two tasks, we pre-train the feature extractor on three real-world WSI datasets: CAMELYON16, TCGA-NSCLC, and TCGA-BRCA, and then evaluate its performance on these tasks. The superior performance of SimMIL further demonstrates its effectiveness. For prognostic prediction, we pre-train the feature extractor on three WSI datasets: TCGA-BLCA, TCGA-LUAD, and TCGA-LUSC, and SimMIL demonstrates exceptional performance in this task as well. Additionally, we highlight the compatibility and scalability of SimMIL by integrating it into state-of-the-art self-supervised learning (SSL) methods and pre-training using merged datasets in benign-malignant classification and cancer subtyping tasks to leverage the extensibility of the datasets in these tasks. We conduct comprehensive ablation studies to explore the key components of SimMIL, providing insights into its design and functionality.

\section{Related Work}
\subsection{Multi-instance Learning for WSI Analysis}
As a \emph{de facto} paradigm for WSI analysis, multi-instance learning can be typically divided into the instance-level and bag-level MIL methods, where the main difference is the way they encode each instance.
\textbf{Instance-level MIL} represents each instance as a score, and a bag score results from the aggregation of the corresponding instances. Most instance-level MIL methods can be summarized into an Expectation-Maximization (EM) algorithmic framework: selecting instances in the E-step and training the model using the selected instances in the M-step~\cite{patchcnn}. Among them, SimpleMIL~\cite{SimpleMIL} is a classic baseline, which directly selects all instances, and follow-up works propose various instance selection strategies~\cite{patchcnn, wang2019weakly, campanella2019clinical, RCE, lin2022interventional}.
\textbf{Bag-level MIL} follows a two-stage modeling approach: encoding each instance as an embedding vector in the first stage and aggregating instances into bag-level representation for MIL tasks in the second stage. 
These attention-based aggregation methods~\cite{abmil,dsmil,clam,transmil,dtfdmil,R2T} have exhibited remarkable performance, but most of them underestimated the importance of feature extractors.
\textbf{Hybrid methods} try to incorporate instance-level and bag-level MIL methods into an integrated network~\cite{bimil, lossmil,chikontwe2020topk,EPL,cluster2conquer}, training the feature extractor and feature aggregator in an end-to-end manner.
Technically sound as they are, the application to WSI analysis is non-trivial:
it is computationally infeasible to use all patches in a WSI due to hardware constraints while sampling partial instances may suffer from missing key information.
Our work lies in the bag-level MIL but focuses on improving the representation quality with the help of instance-level MIL.

\subsection{Visual Representation Learning}
\textbf{Fully supervised pre-training} has emerged as a standard approach in computer vision, and its effectiveness has been proved by various downstream recognition tasks, e.g., image classification~\cite{transfg}, object detection~\cite{fasterrcnn}, segmentation~\cite{maskrcnn}, video action recognition~\cite{videocls}, and MIL on WSI~\cite{transmil,dtfdmil,yao2020whole}. 
The most commonly used natural visual datasets include the ImageNet dataset~\cite{ImageNet, imagenet21k} and the Kinetics dataset~\cite{kinetics}.
However, applying these pre-trained models on WSI suffers from the problem of domain gap, while the annotations of WSI are highly dependent on expertise knowledge and thus are far more expensive.
\textbf{Self-supervised pre-training} (SSL) directly learns the knowledge from large-scale data, which has become a promising solution. 
For example, contrastive learning~\cite{SimCLR,MoCoV2,swav,dino,dinov2} and mask image modeling\cite{beit,mae,beitv3,simmim} have attracted a lot of attention in natural images. 
These general SSL methods have been directly applied to the WSI domain~\cite{dehaene2020self,dsmil,lu2020semi,selfpath,uni,virchow,campanella2023computational}, which outperform the ImageNet pre-trained counterpart.
Some works further design the pathological-specific SSL methods~\cite{kang2023benchmarking,sgcl,BSP,xie2020,sslres18_3,csco}, thus learn more pathological-related patterns.
\textbf{Weakly supervised pre-training} turns to the ``free" labels on the internet, achieving the trade-off between data scale and semantic information~\cite{maewsp,swag}.
They now attract more attention with the development of vision-and-language foundation models~\cite{clip,align}.
For WSI analysis, there are some pioneering works in this line~\cite{ikezogwo2023quilt,plip,conch}, focusing on creating large-scale pathological image and text pairs. 
Instead, we focus on designing the weakly supervised pre-training scheme for MIL using the naturally available weak labels.
\subsection{Pretext Tasks}
Identifying the right pretext task is of vital importance for downstream tasks~\cite{CLsurvey}.
A wide range of pretext tasks have been explored, including rotation prediction~\cite{rotation}, colorization~\cite{colorization}, context auto-encoders~\cite{autoencoders}, inpainting~\cite{inpainting}, jigsaw puzzles~\cite{puzzles}, etc. 
Taking contrastive learning as an example, instance discrimination~\cite{npid} is the default pretext task for classification tasks.
Its variants further explore the local feature and structure information~\cite{densecl,propagateyourself,univip} for dense tasks (e.g., semantic segmentation and object detection), as well as the spatial and temporal information for videos~\cite{videocl,videossl1,videossl2}.
These pretext tasks can be considered as different ways to encode downstream assumptions into pre-training schemes. Our work aims to encode the standard MIL assumption into a weakly supervised pretext task.

\section{Methods}
\label{sec:method}
\subsection{Background}

\noindent\textbf{MIL Formulation.}
The analysis of whole-slide pathological images (WSIs) can be formulated as a multi-instance learning (MIL) problem. Given a dataset $D={X_1,...,X_K}$, each WSI is considered as a bag $X_i={x_{i1}, x_{i2},..., x_{i{N_i}}}$, and each patch $x_{ij}$ is treated as an instance, where $K$ is the number of bags and $N_i$ is the number of instances in the $i^{th}$ bag. During the training process, only the bag-level label $Y_i$ is available, while the instance-level label $y_{ij}$ is presumed to be unavailable.

In the benign-malignant classification task, instance labels are defined as 0 (benign) or 1 (malignant). In the cancer subtyping task, based on the mutually exclusive assumption~\cite{clam}—that different classes cannot coexist within the same slide—instance labels are defined as 0 (negative) or $i$ (where $i$ is the corresponding bag label). Both tasks adhere to the standard MIL assumption, where the relationship between the bag label and the instance labels can be defined as:
\begin{equation}
Y_{i}= 
\begin{cases}
    0, & \text{iff } \sum_j y_{ij}=0 \\ 
    1, & \text{otherwise } 
\end{cases}
\end{equation}

In the survival prediction task, the bag label can be defined as a risk value:
\begin{equation}
    R_i = \phi(t_i, c_i), \quad R_i \in \mathbb{R}
\end{equation}

In this definition, $t_i$ is the survival time and $c_i$ denotes the censorship, which are used in assessing the comparability of a pair of cases and performing comparisons. $R_i$ serves as the learning target of the regression problem. Furthermore, we define the label of the instance $r_{ij}$ in the $i$-th bag:
\begin{equation}
    \forall j \in \{1, \ldots, K\}, \quad r_{ij} \in \mathbb{R}, \quad r_{ij} \ge 0
\end{equation}
where $K$ is the number of instances in a WSI and benign instances are assigned a label of 0. 
In the survival prediction task, the risk of a bag often depends on multiple risky instances rather than a single instance mentioned above. Therefore, we define the relationship between the bag-level and instance-level labels using an accumulative principle:
\begin{equation}
    R_i = \sum_{j=1,...,K} r_{ij}
\end{equation}

\noindent\textbf{Preliminary experiments.} 
\label{preliminary}
As a preliminary step in our work, we conduct experiments on the NCTCRC dataset~\cite{NCTCRC}, which contains nine classes of pathological image patches from human colorectal cancer (CRC) and other normal tissues. 
We pre-train the models using different strategies on \verb=NCR-CRC-HE-100K= and validate using \verb=CRC-VAL-HE-7K= for four downstream tasks: (1) instance-level linear probing, (2) fine-tuning, (3) weakly supervised bag classification with max-pooling, and (4) mean-pooling.
To construct the bags, we follow the MNIST-BAGS approach~\cite{abmil} to label CRC instances as positive and the remaining instances as negative, constructing the NCTCRC-BAGS.
Please refer to the supplementary materials for more details.

\begin{table}[h!]{
    \caption{Preliminary results (\%) of instance-level linear probing (Acc-LP) and fine-tuning (Acc-FT) and weakly supervised bag classification with max/mean pooling aggregator (Max-/Mean-AUC). 
    }
    \label{results_on_NCTCRC}
    {
    \centering
    \begin{center}
    \begin{tabular}{lcccc}
    \hline
    \multirow{2}{*}{\textbf{\makecell{Scheme}}}  &  
     \multicolumn{2}{c}{ NCTCRC} &
    \multicolumn{2}{c}{ NCTCRC-BAGS} \\
     & Acc-LP& Acc-FT  & Max-AUC  & Mean-AUC \\
    \hline 
     ImageNet & 88.66 & 93.71 & 90.60 & 96.64 \\
     MoCo v2 & \textbf{92.14} & 94.45 & 97.46 & 93.67 \\
     SimCLR  & 91.41 & \textbf{94.62} & 98.37 & 95.39  \\
     SimpleMIL & 74.73 & 88.84 &  \textbf{99.04} & \textbf{98.53} \\
     \hline
    \end{tabular}
    \end{center}
    }
    }
\end{table}

~\cref{results_on_NCTCRC} shows the performance of different pre-training schemes on these downstream tasks. We observe an interesting phenomenon where, although the SimpleMIL scheme is less competitive in instance-level tasks, it achieves the best performance in bag classification. 
Since the linear probing and fine-tuning protocols are widely recognized as gold standards for representation learning~\cite{moco,SimCLR,mae}, it would be expected that these "stronger" feature extractors should result in better performance in bag-level tasks, yet the results suggest otherwise.
This suggests that the representation quality in MIL may not be best evaluated by these gold standards, but rather by the downstream MIL performance. 
Therefore, it is essential to develop pre-training schemes tailored to MIL tasks.





\subsection{SimMIL for WSI analysis}
\label{framework}

\begin{figure}[ht!]
    \normalsize
      \centering
      \includegraphics[width=\textwidth]{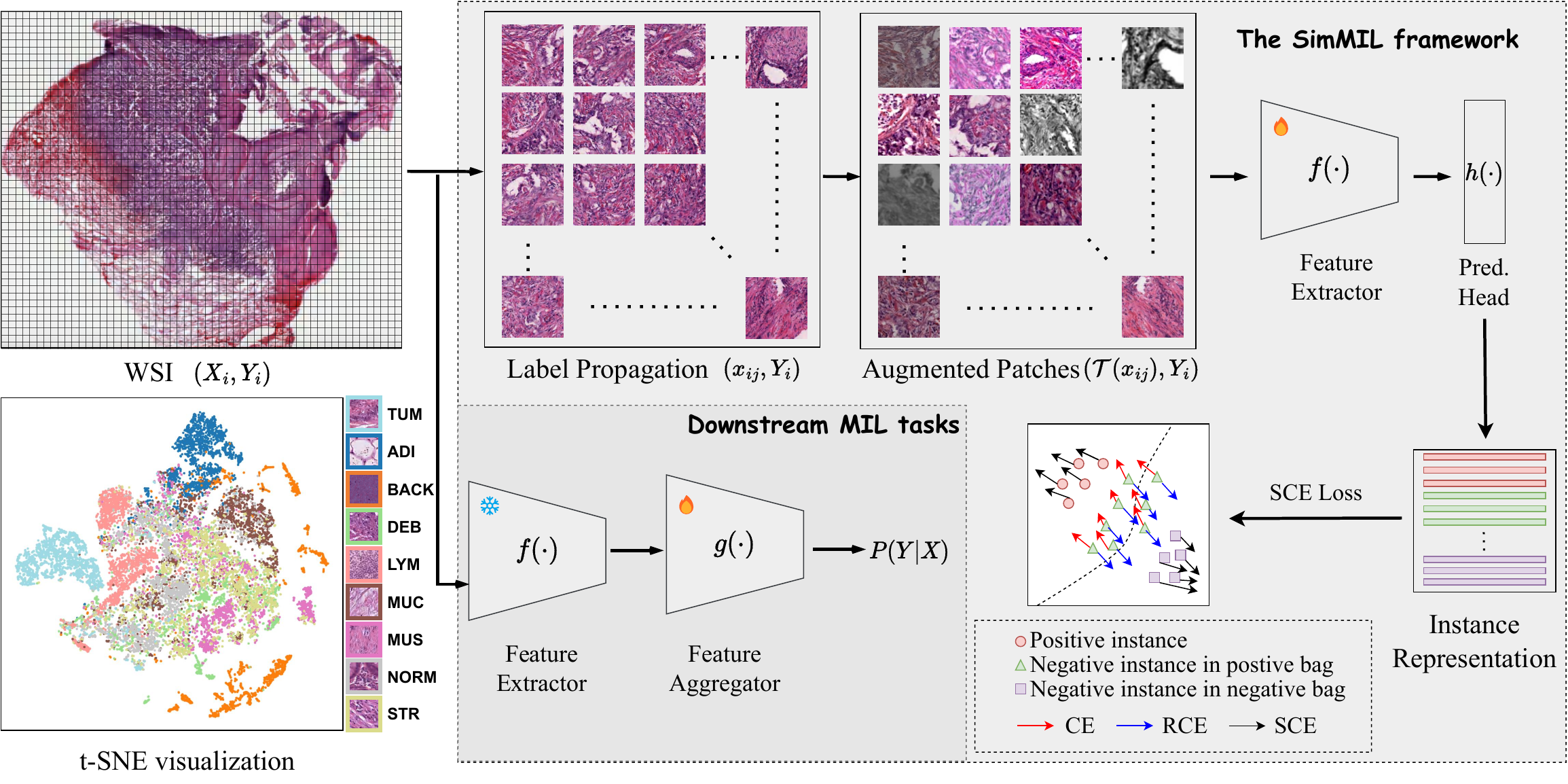}
      \caption{Overview of SimMIL: the pre-training process and downstream MIL tasks. The t-SNE in the bottom left is the visualization result on NCTCRC since the instance-level labels are available.}
      \label{fig:main_fig_framework}
\end{figure}

While SimpleMIL schemes perform well on the synthetic NCTCRC-BAGS, one question that arises is whether they can be directly applied to real-world WSI analysis. Previous works suggest negative outcomes. For example, DSMIL~\cite{dsmil} indicates that SimpleMIL tends to fail when the ratio of positive instances is low (e.g., Camelyon16), while SemiMIL~\cite{wang2019weakly} requires additional instance-level annotations to enable SimpleMIL to learn discriminative patterns. Such additional requirements significantly limit its feasibility in WSI analysis. To eliminate these requirements, we present SimMIL, a simple framework for representation learning in MIL, and explore the key factors that make the SimpleMIL scheme effective.
Besides the feature extractor $f$, the proposed SimMIL consists of several major components:
\begin{itemize}
    \item[1)] \textit{Weakly supervised pre-training scheme.} In MIL settings, only the bag label $Y_i$ is available. The bag label represents a class scalar for benign-malignant classification and cancer subtyping, or the censoring state $c_i$ and survival time $t_i$ for survival prediction. We follow the SimpleMIL~\cite{SimpleMIL} scheme to directly propagate the bag label to the instances within the bag, i.e., $y_{ij} \leftarrow Y_i$, where "$\leftarrow$" denotes assignment, and perform supervised learning on the propagated labels. Under the standard assumption, this scheme can be viewed as an explicit method to encode this assumption. While negative instances are fully supervised, as they share the same label with the bag, instances in positive bags may inevitably encounter the issue of noisy labels, especially when the ratio of positive instances is low. Similarly, under the accumulative assumption in survival prediction, this assignment may incorrectly assign higher risk labels to benign instances. This limitation motivates further exploration of additional components.
    \item[2)] \textit{Data augmentation.} Data augmentation $\mathcal{T}$, a common technique for regularization~\cite{shorten2019survey}, plays an important role in contrastive learning~\cite{SimCLR,MoCoV2} and pathological image analysis~\cite{kang2023benchmarking,IMIL}. To address the noisy label issue in weakly supervised learning, we employ strong data augmentation to increase the diversity of training data, thereby enhancing representation quality. We refer to the data augmentation techniques used in MoCo v2~\cite{MoCoV2}, which include the following strategies: random cropping, random color distortion, random Gaussian blur, and random flipping. 
    \item[3)] \textit{Prediction head.} The prediction head $h$ is designed to transform the feature extractor's output into the prediction target, a standard component of representation learning, such as in contrastive learning~\cite{SimCLR,BYOL} and masked image modeling~\cite{simmim,mae}. Similar to SimSiam~\cite{simsiam}, we hypothesize that $h$ in our framework is specialized to approximate the distribution of augmented images with noisy labels, thus mitigating overfitting in the feature extractor. The implementation of the prediction head follows BYOL~\cite{BYOL}, consisting of two linear layers, an activation layer, and a batch normalization layer.
    \item[4)] \textit{Loss function.} In the benign-malignant classification and cancer subtyping tasks, to effectively counter noisy labels, we integrate Symmetric Cross Entropy (SCE)~\cite{SCE} loss into our framework due to its simplicity and ease of implementation. 
    Put these components together, the overall loss function is:
\begin{equation}
    L_{sce} = -\sum_{i=1}^{|C|}\alpha f(\mathcal{T}(x_i))\log({Y_i}) + \beta {Y_i}\log(f(\mathcal{T}(x_i)))
    \label{sce}
\end{equation}
where $x_i$ is the instance from the $i^{th}$ bag, ${Y_i}$ is the bag label, $\alpha$ and $\beta$ are hyper-parameters controlling the contribution of the two loss components, and $|C|$ is the number of classes. It is worth noting that we directly use CE loss for tumor subtyping tasks to expedite training.

For the survival prediction task, deviating from the approach in~\cite{MCAT}, where regression problems are transformed into classification tasks, we formulate a supervised contrastive learning task using a ranking loss function~\cite{ranking}. The loss function is:
\begin{equation}
    L_{rank} = -\sum_{(x_a,x_b)} \Phi(f(\mathcal{T}(x_a))- f(\mathcal{T}(x_b)))
\end{equation}
 where $x_a$ and $x_b$ are two comparable instances from one batch, with $x_a$ representing a higher risk. $\Phi$ is an indicator approximation function, for which we use the sigmoid function by default.
\end{itemize}

\subsection{Evaluation protocols}
As pointed out in the preliminary experiment, instance-level linear probing and fine-tuning may not adequately evaluate the representation quality for MIL tasks. Instead, it is more appropriate to evaluate using the performance of MIL downstream tasks. Two settings are adopted to align with traditional evaluation protocols:
\begin{itemize}
    \item \textbf{Linear MIL probing:} A linear bag classifier is trained with a non-parametric aggregator (e.g., max-/mean-pooling) over the frozen feature extractor. This approach allows us to evaluate the representation quality likewise the linear probing.
    \item \textbf{Two-stage bag MIL:} A learnable feature extractor is trained over the frozen feature extractor since end-to-end training is computationally infeasible in MIL on WSIs.
\end{itemize}

\subsection{Discussion and analysis}
\paragraph{\textit{Pre}-pretraining scheme}
Inspired by the recent \textit{pre}-pretraining scheme~\cite{prepre}, we propose to further combine SimMIL with self-supervised learning (SSL) approaches to explore SimMIL's compatibility. This experiment is conducted on the benign-malignant classification and cancer subtyping tasks to ensure reliable and robust results. Specifically, we first initialize the feature extractor with state-of-the-art SSL approaches (e.g., CTransPath~\cite{CTrans} and HIPT~\cite{HIPT}), which are specifically designed for WSIs. We then continue to pre-train the feature extractor using the SimMIL scheme. Such a \textit{pre}-pretraining scheme can also be considered a form of iterative training~\cite{taskspecificFT,selfpaceIS,iterativePL}. The key objective of this experiment is to determine whether improved performance can be achieved in a computationally efficient way.


\paragraph{Scaling laws}
Empirical studies show that pre-trained models rely on the scaling of model size, computational resources, and data scale~\cite{mimscaling,clipscaling}. We also conduct a scaling laws study for the MIL pre-training scheme on benign-malignant classification and cancer subtyping, given the ease with which the datasets for these two tasks can be merged and expanded. Our motivation stems from the fact that, according to the WHO classification system, there are dozens of organ sites and hundreds of tumor types, making it impractical to apply SimMIL to every situation. Instead, we can explore scaling laws by merging various datasets and applying weakly-supervised learning for multi-class classification. We can reformulate~\cref{sce} as:
\begin{equation}
    L_{sce} = -\sum_{i=1}^{\hat{|C|}}\alpha f(\mathcal{T}(x_i))\log({Y_i}) + \beta {Y_i}\log(f(\mathcal{T}(x_i)))
    \label{sce_equation}
\end{equation}
where $\hat{|C|}= |\bigcup_{j=1}^{n}C_j|$ and $C_j$ is the class set of $j^{th}$ dataset.

\paragraph{Intuitive understanding}
In the bottom left of~\cref{fig:main_fig_framework}, we visualize the instance features of the SimMIL feature extractor using t-SNE~\cite{tsne}: only the positive instances of the \textit{TUM} class can be separated from others, while other classes cannot be distinguished from one another.
This inductive bias explains the low instance-level performance but strong bag-level MIL performance. On the one hand, bag classification under the standard MIL assumption is about identifying bags that contain \textit{TUM} instances, meaning the bias of the feature extractor aligns precisely with the MIL assumption. 
On the other hand, SimMIL suppresses the interference of negative instances by treating them collectively, which aids the optimization of the aggregator.

\section{Experiments and results}
\label{sec:experiments}

\paragraph{Dataset}
For the benign-malignant classification and cancer subtyping tasks, we conduct experiments on three real-world WSI datasets:
\textbf{Camelyon16}~\cite{CAMELYON16} contains 399 WSIs of  breast cancer. Following~\cite{IMIL}, we extract $\sim2.8$ million patches from 2 tissue classes (metastases vs. normal) at 20 $\times$ magnification. 
\textbf{TCGA-NSCLC} contains two subtypes in lung cancer, Lung Squamous Cell Carcinoma and Lung Adenocarcinoma, with a total of 1,054 WSIs. We directly use the patches released by~\cite{dsmil}, which are $\sim~4.0$ million $224~\times224$ patches at $20~\times$ magnification. For the second stage of HIPT ($\text{ViT}_{4096}-256$), we follow CLAM~\cite{clam} to extract 56,104 $4096\times4096$ regions at 20$\times$ magnification. 
\textbf{TCGA-BRCA}~\cite{TCGA} contains two subtypes in lung cancer, Invasive Ductal and Invasive Lobular Carcinoma, with a total of 1,134 WSIs. Following~\cite{clam}, we extract roughly $\sim3.2$ million $256\times256$ patches and 46,011 $4096\times4096$ regions at 20$\times$ magnification.
Both TCGA-NSCLC and TCGA-BRCA are from The Genome Cancer Atlas (TCGA) project. For Camelyon16, we follow the official split of 270 training images and 129 test images. For TCGA-NSCLC and TCGA-BRCA, we randomly partition them into a training set ($80\%$ cases) and test set (20$\%$ cases) following~\cite{clam}. 

For the survival prediction task, we also conduct experiments on three WSI datasets: \textbf{TCGA-BLCA} contains 375 diagnostic WSIs of Bladder Urothelial Carcinoma. We extract $\sim1.7$ million $256\times256$ patches at 20$\times$ magnification.
\textbf{TCGA-LUAD} and \textbf{TCGA-LUSC}, two subtypes of TCGA-NSCLC, contain 446 and 452 diagnostic WSIs. We extract approximately 1.8 million and 2.1 million 256×256 patches at 20× magnification from these datasets. Following~\cite{clam}, we perform the extraction and divide the datasets into 5-fold cross-validation as outlined in~\cite{R2T}.

\paragraph{Prior arts}
We select the following pre-training methods for comparison:
1) \textbf{ImageNet} pre-trained ResNet18 is the most widely-used feature extractor and we use the version released by Pytorch. 
2) For SSL approaches, we utilize the released models from ConCL~\cite{ConCL} trained for 800 epochs on \verb=NCR-CRC-HE-100K= by two schemes: \textbf{MoCo v2} and \textbf{SimCLR}. 
3) We also compare with some approaches using stronger backbones. \textbf{CLIP} pre-trains on 400 million natural (image, text) pairs with contrastive learning, using ResNet50 as the backbone. \textbf{PLIP}~\cite{plip} fine-tunes CLIP on 208,414 pathology (image, text) pairs with contrastive learning, using ViT-B as the backbone. \textbf{CTransPath}~\cite{CTrans} uses a hybrid CNN and Transformer architecture, pre-trained on 15 million patches from WSIs in TCGA~\cite{TCGA} and PAIP\footnote{\url{http://www.wisepaip.org/paip/}}. \textbf{HIPT} pre-trains a hierarchical Transformer with 408,218 4096$\times$4096 regions and 104 millon 256$\times$256 patches at 20$\times$ magnification from TCGA. 

\paragraph{MIL aggregators}
We verify the performance of pre-trained feature extractors on \textbf{max-} and \textbf{mean-pooling}~\cite{pooling} for linear probing and three attention-based networks (\textbf{ABMIL}~\cite{abmil}, \textbf{DSMIL}~\cite{dsmil},  \textbf{CLAM-SB}~\cite{clam}), \textbf{TransMIL}~\cite{transmil} and \textbf{DTFD-MIL}~\cite{dtfdmil} for two-stage bag MIL. 
Refer to supplementary for more details.

\subsection{Benign-malignant Classification and Cancer Subtyping}
\label{main_experiment}

\paragraph{Experiment settings}
For SimMIL pre-training, we train ResNet18 using an SGD optimizer with no weight decay, a momentum of 0.9, and a batch size of 256 on four GPUs.
As described in~\cref{sec:method}, we apply SCE loss for classification on the Camelyon16 dataset ($\alpha = 1.0$, $\beta = 1.0$) and CE loss for the subtyping tasks (i.e., TCGA-NSCLC and TCGA-BRCA), with an initial learning rate of 0.001 and a stepwise learning rate scheduler for 200 and 100 epochs, respectively, to avoid overfitting noisy labels.
Given the pre-trained feature extractors, we train the linear classifiers and aggregation networks for 50 epochs using the Adam optimizer and a cosine annealing scheduler.


\paragraph{Results}

\begin{table}[ht!]
    \centering
    \resizebox{\linewidth}{!}{
    \begin{tabular}{ccccccccc}
    \toprule
    \multirow{2}{*}{\textbf{Agg.}} & 
    \multirow{2}{*}{\textbf{Method}} &
    \multirow{2}{*}{\textbf{Arch.}} &
    \multicolumn{2}{c}{Camelyon16} & \multicolumn{2}{c}{TCGA-NSCLC} & \multicolumn{2}{c}{TCGA-BRCA} \\
    \cmidrule(lr){4-5}  \cmidrule(lr){6-7} 
    \cmidrule(lr){8-9}  
     &  &  & Acc & AUC & Acc & AUC & Acc & AUC \\ 
     \midrule
    \multirow{9}{*}[2ex]{\rotatebox{90}{\textbf{Max Pooling}}} & ImageNet & ResNet18 & 53.75 \footnotesize{$\pm$ 2.22} & 54.33 \footnotesize{$\pm$ 2.81} & 77.94 \footnotesize{$\pm$ 0.60} & 86.22 \footnotesize{$\pm$ 0.05} & 70.30 \footnotesize{$\pm$ 1.31} & 79.46 \footnotesize{$\pm$ 0.41} \\
    & MoCo v2 & ResNet18 & 59.43 \footnotesize{$\pm$ 1.59}  & 62.08 \footnotesize{$\pm$ 0.47}  & 77.30 \footnotesize{$\pm$ 0.45} & 82.99 \footnotesize{$\pm$ 0.17} & 63.84 \footnotesize{$\pm$ 2.98}  & 67.34 \footnotesize{$\pm$ 0.14}  \\
    & SimCLR & ResNet18 & \underline{68.74 \footnotesize{$\pm$ 1.46}}  & 60.61 \footnotesize{$\pm$ 1.33}  & 77.46 \footnotesize{$\pm$ 3.26}  & 89.06 \footnotesize{$\pm$ 0.13} & 76.36 \footnotesize{$\pm$ 1.31} & 78.58 \footnotesize{$\pm$ 0.07} \\
    & CLIP & ResNet50 & 67.70 \footnotesize{$\pm$ 1.32} & 61.49 \footnotesize{$\pm$ 0.23} & 63.17 \footnotesize{$\pm$ 2.50} & 73.67 \footnotesize{$\pm$ 0.06} & 57.78 \footnotesize{$\pm$ 1.43} & 61.09 \footnotesize{$\pm$ 0.15} \\
    & SRCL & CTransPath & 66.92 \footnotesize{$\pm$ 1.93} & \underline{66.03 \footnotesize{$\pm$ 1.45}} & 82.06 \footnotesize{$\pm$ 0.81} & 90.38 \footnotesize{$\pm$ 0.07} & \underline{78.59 \footnotesize{$\pm$ 0.29}} & 81.24 \footnotesize{$\pm$ 0.18} \\
    & DINO & HIPT & - & - & \underline{82.41 \footnotesize{$\pm$ 1.59}} & \underline{91.26 \footnotesize{$\pm$ 0.05}} & 73.58 \footnotesize{$\pm$ 3.20} & \underline{86.88 \footnotesize{$\pm$ 0.22}} \\
    & SimMIL\footnotesize{(ours)} & ResNet18 & \textbf{79.33 \footnotesize{$\pm$ 0.97}} & \textbf{78.29 \footnotesize{$\pm$ 1.10}} & \textbf{ 88.57 \footnotesize{$\pm$ 0.78 }} & \textbf{95.87 \footnotesize{$\pm$ 0.03}} & \textbf{87.47 \footnotesize{$\pm$ 0.29}} & \textbf{90.99 \footnotesize{$\pm$ 0.09}} \\
    \midrule
    \multirow{9}{*}[2ex]{\rotatebox{90}{\textbf{Mean Pooling}}} & ImageNet & ResNet18 & 63.57 \footnotesize{$\pm$ 0.00} & 49.32 \footnotesize{$\pm$ 0.37} & 78.57 \footnotesize{$\pm$ 0.67} & 84.36 \footnotesize{$\pm$ 0.02} & 63.84 \footnotesize{$\pm$ 3.51} & 78.87 \footnotesize{$\pm$ 0.29} \\
    & MoCo v2 & ResNet18 & 44.96 \footnotesize{$\pm$ 0.00 }  & 44.40 \footnotesize{$\pm$ 0.41 }  & 71.59 \footnotesize{$\pm$ 0.22} & 75.62 \footnotesize{$\pm$ 0.01} & 67.27 \footnotesize{$\pm$ 0.99}  & 58.43 \footnotesize{$\pm$ 0.19}  \\
    & SimCLR & ResNet18 & 65.63 \footnotesize{$\pm$ 0.36}  & 45.78 \footnotesize{$\pm$ 0.12}  & 83.02 \footnotesize{$\pm$ 
    0.22}  & 88.03 \footnotesize{$\pm$ 0.06} & 73.54 \footnotesize{$\pm$ 1.43} & 76.74 \footnotesize{$\pm$ 0.08} \\
    & CLIP & ResNet50 & 59.43 \footnotesize{$\pm$ 0.74} & 56.57 \footnotesize{$\pm$ 0.07} & 72.70 \footnotesize{$\pm$ 0.45} & 77.79 \footnotesize{$\pm$ 0.02} & 63.84 \footnotesize{$\pm$ 1.43} & 63.94 \footnotesize{$\pm$ 0.03} \\
    & SRCL & CTransPath & 65.63 \footnotesize{$\pm$ 0.36} & 38.72 \footnotesize{$\pm$ 0.16} & \underline{86.03 \footnotesize{$\pm$ 0.23}} & 91.11 \footnotesize{$\pm$ 0.02} & \underline{77.78 \footnotesize{$\pm$ 0.28}} & 84.59 \footnotesize{$\pm$ 0.05} \\
    & DINO & HIPT & - & - & 84.26 \footnotesize{$\pm$ 0.26} & \underline{91.39 \footnotesize{$\pm$ 0.11}} & 63.21 \footnotesize{$\pm$ 3.20} & \underline{87.38 \footnotesize{$\pm$ 0.17}} \\
    & SimMIL\footnotesize{(ours)} & ResNet18 & \textbf{74.68 \footnotesize{$\pm$ 0.36}} & \textbf{59.09 \footnotesize{$\pm$ 0.28}} & \textbf{86.35 \footnotesize{$\pm$ 0.23}} & \textbf{92.82 \footnotesize{$\pm$ 0.04}} & \textbf{85.05 \footnotesize{$\pm$ 0.76}} & \textbf{87.67 \footnotesize{$\pm$ 0.06}} \\
    \midrule
    \multirow{9}{*}[2ex]{\rotatebox{90}{\textbf{ABMIL}}} & ImageNet & ResNet18 & 80.36 \footnotesize{$\pm$ 0.37} & \underline{78.77 \footnotesize{$\pm$ 1.56}} & 81.59 \footnotesize{$\pm$ 0.22} & 87.80 \footnotesize{$\pm$ 1.45} & 69.09 \footnotesize{$\pm$ 3.01} & 64.93 \footnotesize{$\pm$ 2.59} \\
    & MoCo v2 & ResNet18 & 73.90 \footnotesize{$\pm$ 0.96}  & 78.19 \footnotesize{$\pm$ 0.44}  & 84.29 \footnotesize{$\pm$ 0.67} & 87.45 \footnotesize{$\pm$ 0.22} & 52.12 \footnotesize{$\pm$ 0.86}  & 58.12 \footnotesize{$\pm$ 0.33}  \\
    & SimCLR & ResNet18 & 76.75 \footnotesize{$\pm$ 2.29}  & 74.24 \footnotesize{$\pm$ 1.88}  & 84.92 \footnotesize{$\pm$ 
    1.75}  & 90.54 \footnotesize{$\pm$ 0.69} & 71.31 \footnotesize{$\pm$ 2.06} & 80.26 \footnotesize{$\pm$ 1.14} \\
    & CLIP & ResNet50 & 63.57 \footnotesize{$\pm$ 0.00} & 63.83 \footnotesize{$\pm$ 0.49} & 83.49 \footnotesize{$\pm$ 0.90} & 88.35 \footnotesize{$\pm$ 0.54} & 75.15 \footnotesize{$\pm$ 0.85} & 69.28 \footnotesize{$\pm$ 0.48} \\
    & SRCL & CTransPath & \textbf{83.46 \footnotesize{$\pm$ 0.36}} & 78.28 \footnotesize{$\pm$ 0.21} & \underline{89.21 \footnotesize{$\pm$ 1.62}} & \underline{94.27 \footnotesize{$\pm$ 0.30}} & 84.44 \footnotesize{$\pm$ 0.29} & 78.45 \footnotesize{$\pm$ 1.01} \\
    & DINO & HIPT & - & - & 86.30 \footnotesize{$\pm$ 0.69} & 93.26 \footnotesize{$\pm$ 0.29} & \textbf{87.20 \footnotesize{$\pm$ 1.31}} & \underline{82.94 \footnotesize{$\pm$ 1.89}} \\
    & SimMIL\footnotesize{(ours)} & ResNet18 & \underline{81.39 \footnotesize{$\pm$ 1.68}} & \textbf{85.21 \footnotesize{$\pm$ 1.36}} & \textbf{90.32 \footnotesize{$\pm$ 1.36}} & \textbf{96.17 \footnotesize{$\pm$ 0.53}} & \underline{84.24 \footnotesize{$\pm$ 1.49}} & \textbf{91.49 \footnotesize{$\pm$ 0.40}} \\
    \midrule
    \multirow{9}{*}[2ex]{\rotatebox{90}{\textbf{DSMIL}}} & ImageNet & ResNet18 & 74.42 \footnotesize{$\pm$ 3.35} & 68.35 \footnotesize{$\pm$ 2.95} & 75.87 \footnotesize{$\pm$ 3.12} & 85.78 \footnotesize{$\pm$ 1.73} & 57.98 \footnotesize{$\pm$ 2.44} & 72.26 \footnotesize{$\pm$ 1.69} \\
    & MoCo v2 & ResNet18 & 61.50 \footnotesize{$\pm$ 0.96} & 55.80 \footnotesize{$\pm$ 0.71} & 72.37 \footnotesize{$\pm$ 2.36} & 84.77 \footnotesize{$\pm$ 0.17} & 50.51 \footnotesize{$\pm$ 1.03} & 59.62 \footnotesize{$\pm$ 0.12} \\
    & SimCLR & ResNet18 & \underline{75.97 \footnotesize{$\pm$ 1.90}} & 73.28 \footnotesize{$\pm$ 0.44} & 83.33 \footnotesize{$\pm$ 1.56} & 90.68 \footnotesize{$\pm$ 0.65} & 64.85 \footnotesize{$\pm$ 1.78} & 79.04 \footnotesize{$\pm$ 2.80} \\
    & CLIP & ResNet50 & 63.31 \footnotesize{$\pm$ 0.73} & 58.47 \footnotesize{$\pm$ 0.47} & 74.60 \footnotesize{$\pm$ 0.81} & 83.16 \footnotesize{$\pm$ 0.23} & 67.88 \footnotesize{$\pm$ 0.49} & 71.19 \footnotesize{$\pm$ 0.14} \\
    & SRCL & CTransPath & 70.54 \footnotesize{$\pm$ 0.96} & \underline{79.83 \footnotesize{$\pm$ 0.59}} & \underline{88.73 \footnotesize{$\pm$ 0.59}} & \underline{95.17 \footnotesize{$\pm$ 0.09}} & \textbf{77.98 \footnotesize{$\pm$ 1.03}} & \textbf{90.96 \footnotesize{$\pm$ 0.40}} \\
    & DINO & HIPT & - & - & 84.82 \footnotesize{$\pm$ 0.69} & 93.91 \footnotesize{$\pm$ 0.16} & 75.61 \footnotesize{$\pm$ 1.80} & 88.37 \footnotesize{$\pm$ 0.50} \\
    & SimMIL\footnotesize{(ours)} & ResNet18 & \textbf{79.59 \footnotesize{$\pm$ 1.32}} & \textbf{82.10 \footnotesize{$\pm$ 1.53}} & \textbf{89.05 \footnotesize{$\pm$ 1.17}} & \textbf{95.40 \footnotesize{$\pm$ 0.12}} & \underline{76.16 \footnotesize{$\pm$ 1.51}} & \underline{89.86 \footnotesize{$\pm$ 0.44}} \\
    \midrule
    \multirow{9}{*}[2ex]{\rotatebox{90}{\textbf{CLAM-SB}}} & ImageNet & ResNet18 & 77.26 \footnotesize{$\pm$ 2.93} & 76.04 \footnotesize{$\pm$ 2.21} & 84.29 \footnotesize{$\pm$ 0.67} & 90.30 \footnotesize{$\pm$ 0.43} & 77.37 \footnotesize{$\pm$ 1.87} & 80.11 \footnotesize{$\pm$ 2.54} \\
    & MoCo v2 & ResNet18 & 78.82 \footnotesize{$\pm$ 0.38} & 78.14 \footnotesize{$\pm$ 1.05} & 86.19 \footnotesize{$\pm$ 1.56} & 91.91 \footnotesize{$\pm$ 1.94} & \underline{83.03 \footnotesize{$\pm$ 1.78}} & 87.08 \footnotesize{$\pm$ 1.05} \\
    & SimCLR & ResNet18 & 75.97 \footnotesize{$\pm$ 2.28} & 74.34 \footnotesize{$\pm$ 1.24} & 86.35 \footnotesize{$\pm$ 0.45} & 91.99 \footnotesize{$\pm$ 0.67} & 81.82 \footnotesize{$\pm$ 2.27} & 79.43 \footnotesize{ $\pm$ 0.37} \\
    & CLIP & ResNet50 & 72.87 \footnotesize{$\pm$ 0.63} & 63.51 \footnotesize{$\pm$ 0.23} & 86.83 \footnotesize{$\pm$ 0.59} & 92.28 \footnotesize{$\pm$ 0.62} & 80.61 \footnotesize{$\pm$ 0.86} & 89.71 \footnotesize{$\pm$ 0.02} \\
    & SRCL & CTransPath & \textbf{84.50 \footnotesize{$\pm$ 0.01}} & \underline{80.84 \footnotesize{$\pm$ 1.28}} & 89.05 \footnotesize{$\pm$ 0.67} & 95.03 \footnotesize{$\pm$ 0.19} & 81.21 \footnotesize{$\pm$ 1.78} & 83.89 \footnotesize{$\pm$ 0.46} \\
    & DINO & HIPT & - & - &\underline{90.19 \footnotesize{$\pm$ 0.26}} & \underline{95.95 \footnotesize{$\pm$ 0.38}} & 80.49 \footnotesize{$\pm$ 2.99} & \textbf{92.23 \footnotesize{$\pm$ 0.52}} \\
    & SimMIL\footnotesize{(ours)} & ResNet18 & \underline{83.47 \footnotesize{$\pm$ 1.46}} & \textbf{84.35 \footnotesize{$\pm$ 1.76}} & \textbf{93.65 \footnotesize{$\pm$ 0.81}} & \textbf{97.19 \footnotesize{$\pm$ 0.83}} & \textbf{88.48 \footnotesize{$\pm$ 0.99}} & \underline{92.08 \footnotesize{$\pm$ 0.42}} \\
    \bottomrule
    \end{tabular}}
    \caption{Reuslts (\%) of Bag-level classification on Camelyon16, TCGA-NSCLC and TCGA-BRCA. Acc and AUC are reported.}
    \label{benchmarking_on_C16_TCGA_BRCA}
\end{table}

~\cref{benchmarking_on_C16_TCGA_BRCA} shows the results of linear MIL probing and two-stage bag MIL on Camelyon16, TCGA-NSCLC and TCGA-BRCA. 
For each setting, we report the average and standard deviation of Accuracy and AUC across three experiments with different random seeds. 
Using the same ResNet18 backbone, SimMIL outperforms ImageNet pre-training and SSL approaches across all aggregation networks. Compared to stronger backbones, such as CTransPath and HIPT, SimMIL achieves the best results on TCGA-NSCLC and comparable, if not superior, results on Camelyon16 and TCGA-BRCA.
Overall, the results demonstrate the effectiveness of SimMIL.

\subsection{Survival Prediction}

\paragraph{Experiment settings}
For pre-training in survival prediction, we train ResNet18 using an SGD optimizer with no weight decay, a momentum of 0.9, and a batch size of 256 on four GPUs. We set the learning rate to 0.001 and apply a cosine annealing scheduler over 100 epochs for the ranking loss. With the pre-trained feature extractors, we follow~\cite{MCAT} to train with an NLL loss on downstream tasks for 30 epochs. More implementation details can be found in the supplementary material.

\begin{table}[ht!]
    \centering
    \resizebox{\linewidth}{!}{
    \begin{tabular}{cccccc}
    \toprule
    \textbf{Agg.} & 
    \textbf{Method} &
    \textbf{Arch.} &
    TCGA-LUAD & TCGA-BLCA & TCGA-LUSC \\
    \midrule
    \multirow{6}{*}[2ex]{\rotatebox{90}{\textbf{ABMIL}}} & ImageNet & ResNet18 & 54.29 \footnotesize{$\pm$ 3.25} & 50.59 \footnotesize{$\pm$ 4.68} & 48.43 \footnotesize{$\pm$ 3.08} \\
    & MoCo v2 & ResNet18 & \underline{56.52  \footnotesize{$\pm$ 6.45}}  & \textbf{58.56  \footnotesize{$\pm$ 4.04}}  &  \underline{59.50  \footnotesize{$\pm$ 2.75  }}  \\
    & SimCLR & ResNet18 & 56.22  \footnotesize{$\pm$ 4.80 }  & 54.06  \footnotesize{$\pm$ 5.36 }  & 48.59 \footnotesize{$\pm$ 3.24 } \\
    & PLIP & ViT-B & 50.20  \footnotesize{$\pm$ 5.53 } & 51.74  \footnotesize{$\pm$ 3.54} & 50.73  \footnotesize{$\pm$ 5.37 } \\
    & SimMIL\footnotesize{(ours)} & ResNet18 & \textbf{58.62 \footnotesize{$\pm$ 3.65}} & \underline{57.49 \footnotesize{$\pm$ 7.99}} & \textbf{ 59.66 \footnotesize{$\pm$ 6.77 }} \\
    \midrule
    \multirow{6}{*}[2ex]{\rotatebox{90}{\textbf{CLAM-SB}}} & ImageNet & ResNet18 & 57.48 \footnotesize{$\pm$ 1.84 } & 46.06 \footnotesize{$\pm$ 6.36} & 50.88  \footnotesize{$\pm$ 2.10 } \\
    & MoCo v2 & ResNet18 & \underline{58.23  \footnotesize{$\pm$ 7.07}} & 54.36  \footnotesize{$\pm$ 2.72} & \underline{59.41 \footnotesize{$\pm$ 4.08 }} \\
    & SimCLR & ResNet18 & 56.74  \footnotesize{$\pm$ 5.75} & 52.86  \footnotesize{$\pm$ 5.71} &  53.22 \footnotesize{$\pm$ 4.09 } \\
    & PLIP & ViT-B & 53.52  \footnotesize{$\pm$ 4.46} & \underline{54.68  \footnotesize{$\pm$ 2,71}} & 50.74 \footnotesize{$\pm$ 5.43 } \\
    & SimMIL\footnotesize{(ours)} & ResNet18 & \textbf{59.17  \footnotesize{$\pm$ 2.51}} & \textbf{56.52  \footnotesize{$\pm$ 3.69}} & \textbf{61.33  \footnotesize{$\pm$ 2.50 }} \\
    \midrule
    \multirow{6}{*}[2ex]{\rotatebox{90}{\textbf{TransMIL}}} & ImageNet & ResNet18 & \underline{60.20  \footnotesize{$\pm$ 3.73}} & 55.09  \footnotesize{$\pm$ 0.76} & 58.45  \footnotesize{$\pm$ 4.58 } \\
    & MoCo v2 & ResNet18 & 55.20  \footnotesize{$\pm$ 5.84} & 56.67  \footnotesize{$\pm$ 7.11} & \underline{59.69 \footnotesize{$\pm$ 3.89 }} \\
    & SimCLR & ResNet18 & 59.15  \footnotesize{$\pm$ 8.98} & \underline{59.65 \footnotesize{$\pm$ 4.76}} & 56.04  \footnotesize{$\pm$ 5.51 } \\
    & PLIP & ViT-B & 59.17  \footnotesize{$\pm$ 8.34} & 56.50  \footnotesize{$\pm$ 2.74} & 56.92 \footnotesize{$\pm$ 3.39 } \\
    & SimMIL\footnotesize{(ours)} & ResNet18 & \textbf{60.35 \footnotesize{$\pm$ 3.28}} & \textbf{59.94 \footnotesize{$\pm$ 4.67}} & \textbf{ 60.61 \footnotesize{$\pm$ 1.31 }} \\
    \midrule
    \multirow{6}{*}[2ex]{\rotatebox{90}{\textbf{DTFD-MIL}}} & ImageNet & ResNet18 & \underline{60.37 \footnotesize{$\pm$ 5.09}} & 59.77 \footnotesize{$\pm$ 2.29} & 53.03 \footnotesize{$\pm$ 4.85 } \\
    & MoCo v2 & ResNet18 & 58.99 \footnotesize{$\pm$ 3.75} & 57.18 \footnotesize{$\pm$ 4.46} & 47.74 \footnotesize{$\pm$ 2.98 } \\
    & SimCLR & ResNet18 & 60.00 \footnotesize{$\pm$ 5.93} & \underline{60.63 \footnotesize{$\pm$ 5.50}} &  50.92 \footnotesize{$\pm$ 4.10 } \\
    & PLIP & ViT-B & 52.71 \footnotesize{$\pm$ 2.91} & 55.16 \footnotesize{$\pm$ 2.18} & \underline{55.86 \footnotesize{$\pm$ 3.97 }} \\
    & SimMIL\footnotesize{(ours)} & ResNet18 & \textbf{61.96 \footnotesize{$\pm$ 4.15}} & \textbf{61.94 \footnotesize{$\pm$ 6.80}} & \textbf{ 56.59 \footnotesize{$\pm$ 4.45 }} \\
    \bottomrule
    \end{tabular}
    }
    \caption{Reuslts (\%) of Survival prediction on TCGA-LUAD, TCGA-BLCA and TCGA-LUSC. C-index is reported.}
    \label{benchmarking_on_LUAD_BLCA_LUSC}
\end{table}

\paragraph{Results}
~\cref{benchmarking_on_LUAD_BLCA_LUSC} presents the results of two-stage bag MIL on TCGA-LUAD, TCGA-BLCA, and TCGA-LUSC. We report the average and standard deviation of the C-index across experiments using 5-fold cross-validation datasets. In almost all experiments, SimMIL achieves the best downstream results, except for one case (downstream training with ABMIL on TCGA-BLCA), where SimMIL achieves the second-best result. 
In summary, these results validate the effectiveness of the SimMIL framework for survival prediction tasks.

\subsection{Fine-tuning on WSI-specific SSL approaches}

\begin{figure}[ht!]
    \centering
    \begin{subfigure}{0.48\linewidth}
      \centering
      \includegraphics[width=\linewidth]{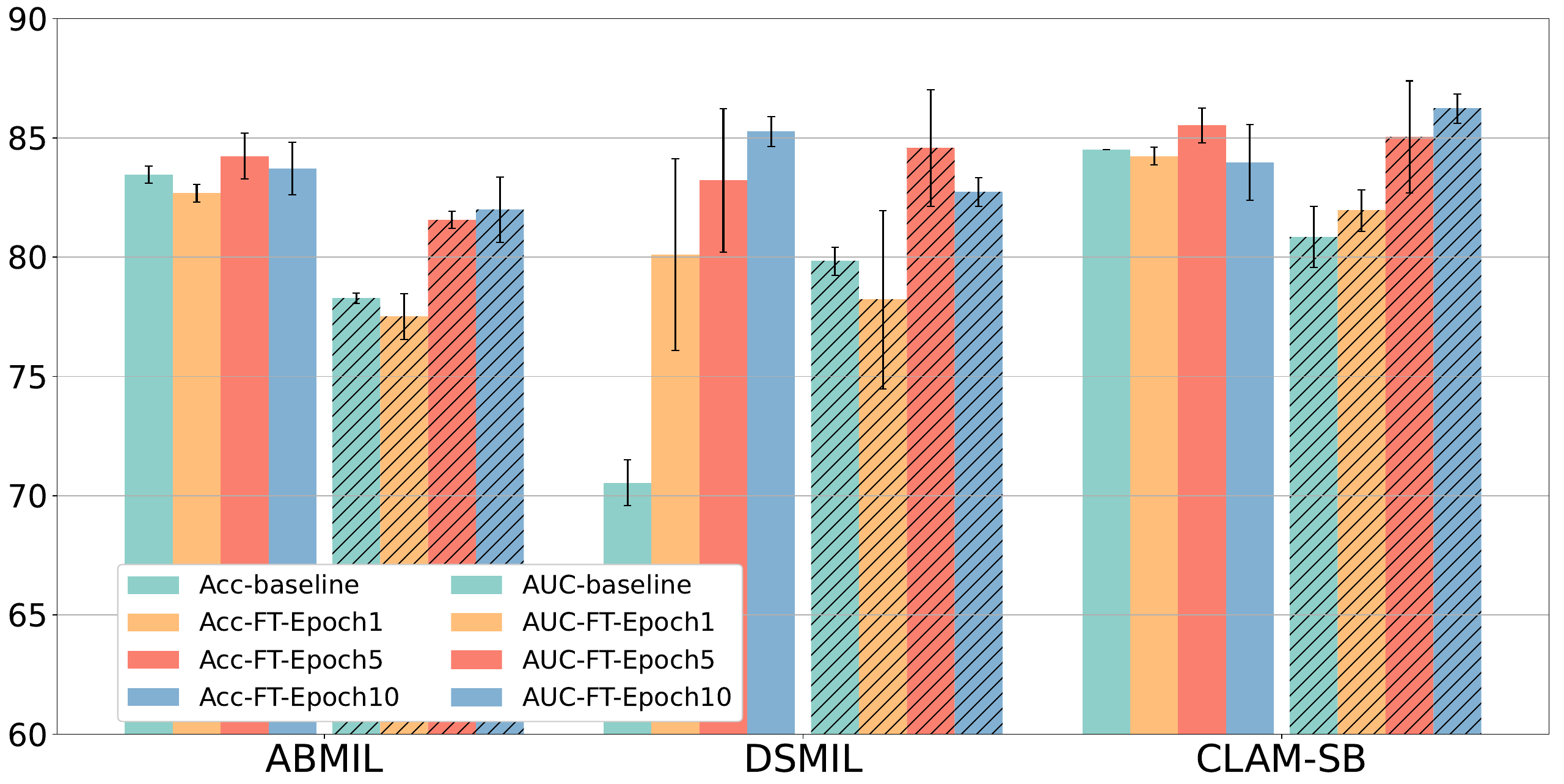}
      \caption{Fine-tuning CTransPath on Camelyon16.}
      \label{C16_CTrans}
    \end{subfigure}
    \hfill
    \begin{subfigure}{0.48\linewidth}
      \centering
      \includegraphics[width=\linewidth]{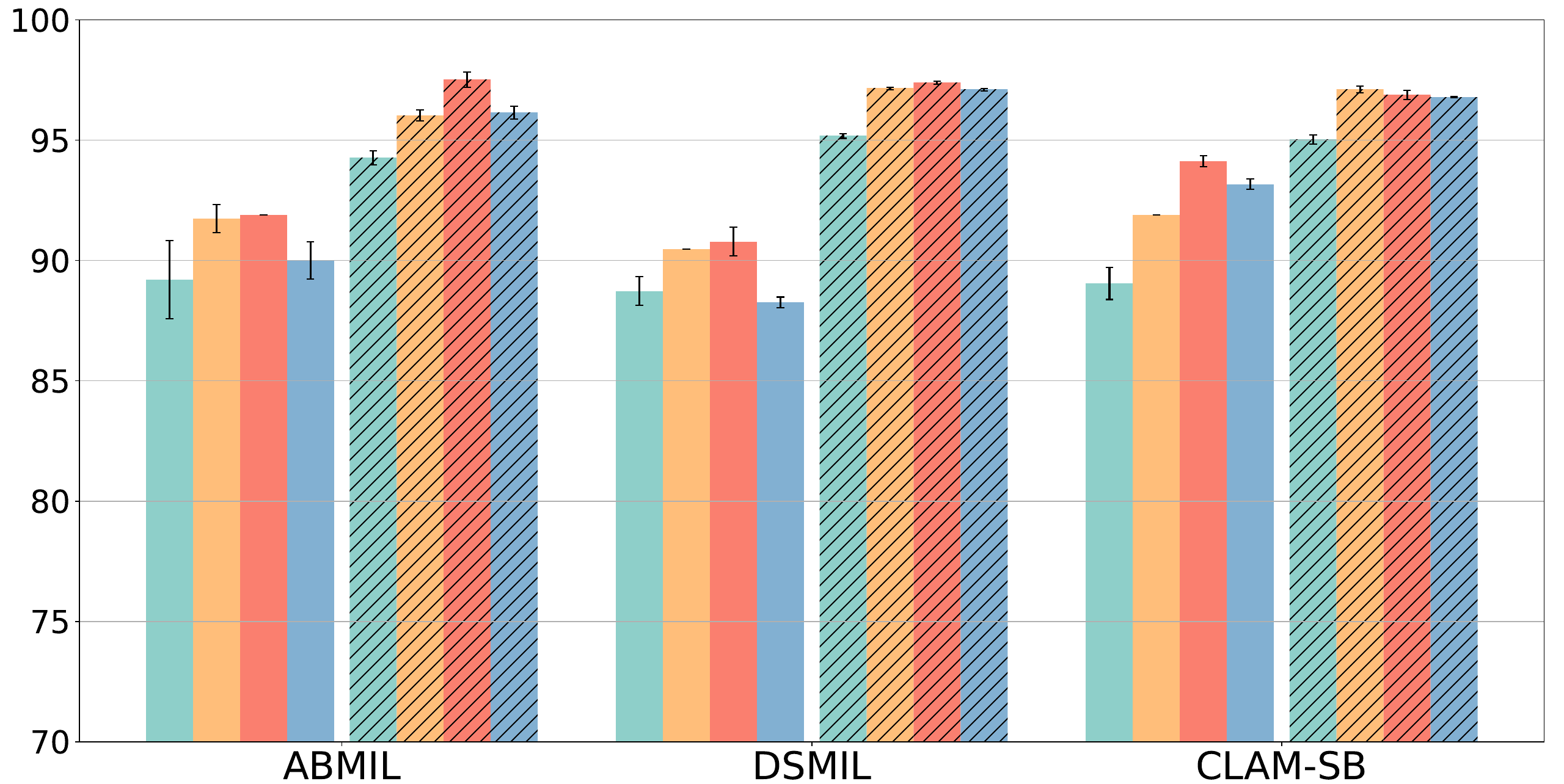}
      \caption{Fine-tuning CTransPath on TCGA-NSCLC.}
      \label{TCGA_CTrans}
    \end{subfigure}
  
    \begin{subfigure}{0.48\linewidth}
      \centering
      \includegraphics[width=\linewidth]{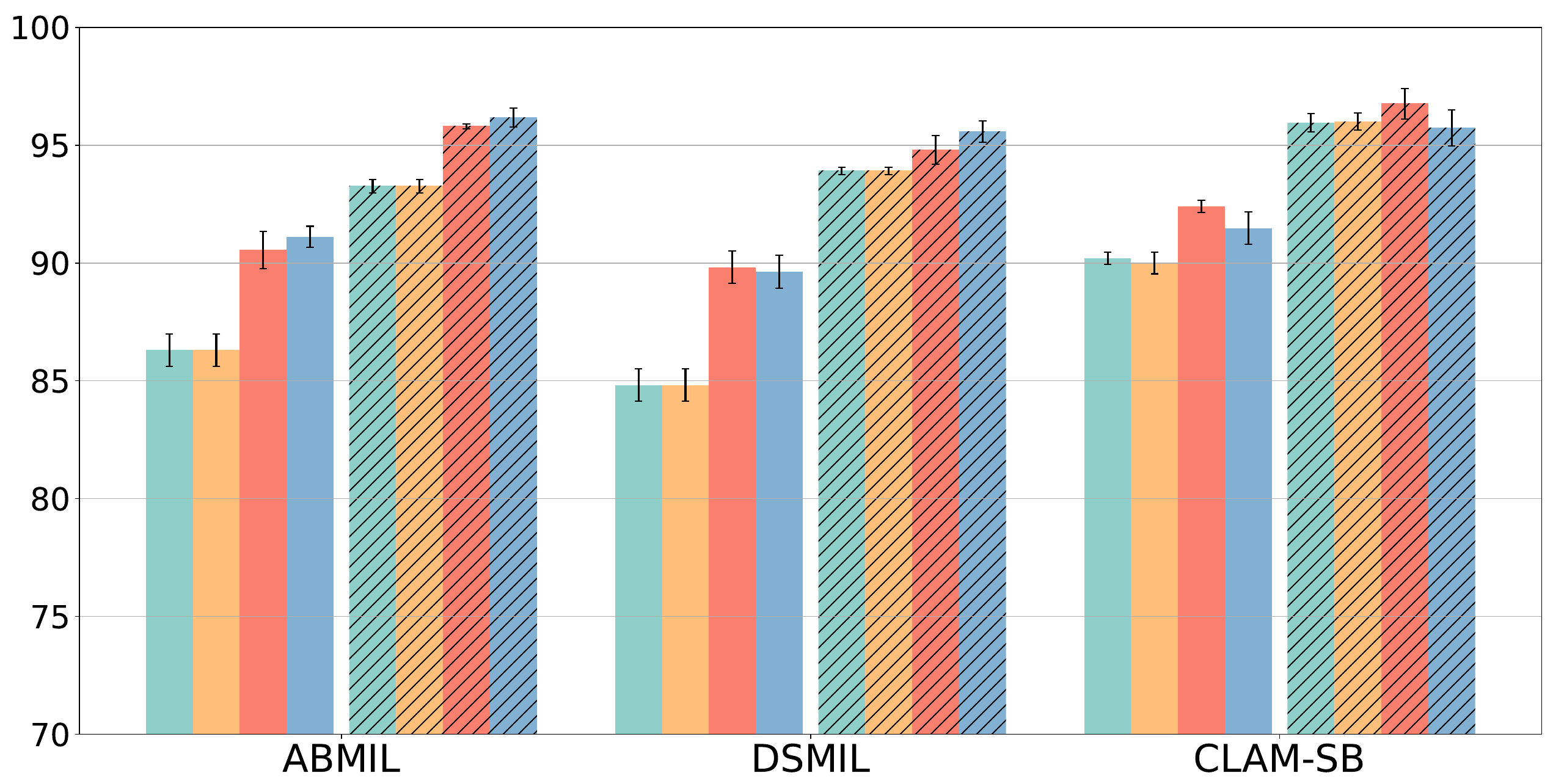}
      \caption{Fine-tuning HIPT on TCGA-NSCLC.}
      \label{TCGA_HIPT}
    \end{subfigure}
    \hfill
    \begin{subfigure}{0.48\linewidth}
      \centering
      \includegraphics[width=\linewidth]{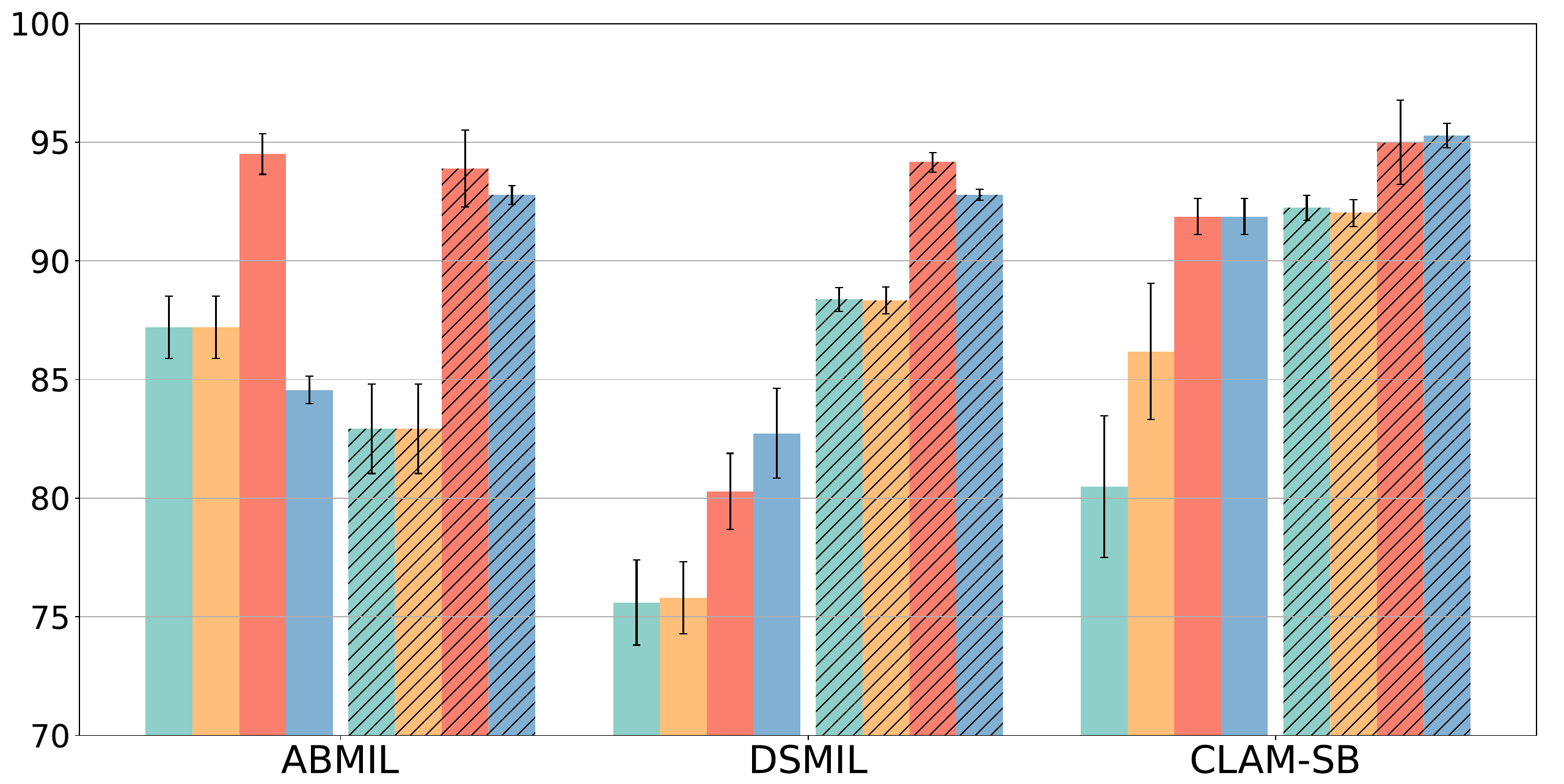}
      \caption{Fine-tuning HIPT on TCGA-BRCA.}
      \label{BRCA_HIPT}
    \end{subfigure}
  \caption{Reuslts (\%) of bag-level classification with feature extractor released and fine-tuned CTransPath on Camelyon16 and TCGA-NSCLC, HIPT on TCGA-NSCLC and TCGA-BRCA. FT-Epoch1 denotes fine-tuning for 1 epoch and analogously for FT-Epoch5 and FT-Epoch10. Acc and AUC are reported by solid-colored and striped bars. Legend is the same for 4 subfigures.}
  \label{fig:finetune}
  \end{figure}

\paragraph{Experiment settings}
To explore the compatibility between SimMIL and SSL approaches, we fine-tune two state-of-the-art SSL models, namely CTransPath and HIPT. CTransPath directly encodes patches with a single-stage transformer, so we train it end-to-end. HIPT, on the other hand, divides the input images into two levels ($256\times256$ and $4096\times4096$) and computes features hierarchically; therefore, we fine-tune the second stage due to the larger input resolution, which introduces less label noise. Both CTransPath and HIPT are fine-tuned using the SGD optimizer and a cosine annealing scheduler with an initial learning rate of $5e-4$. We train these models for 1, 5, and 10 epochs to explore the efficiency and performance trend in fine-tuning. The average and standard deviation of Accuracy and AUC across three experiments are shown in the figure.


\paragraph{Results}
Figure~\ref{fig:finetune} presents the results of fine-tuning experiments. We can observe that 1) Downstream performance on all datasets for both SSL models improves after fine-tuning with our SimMIL model. 2) This improvement can be achieved with only 5 epochs of fine-tuning, and in some settings, even 1 epoch is sufficient, such as fine-tuning CTransPath on TCGA-NSCLC, as shown in~\cref{TCGA_CTrans}. 3) After fine-tuning for 10 epochs, performance in several settings decreases, e.g., fine-tuning HIPT on TCGA-BRCA. We believe that this phenomenon stems from model overfitting. Overall, these results demonstrate SimMIL's compatibility and its ability to enhance other SoTA methods in a highly efficient manner.


\subsection{Scaling Experiments on Merged Datasets}

\begin{table}[ht!]
    \centering
    \resizebox{\linewidth}{!}
{
    \begin{tabular}{cccccccc}
    \toprule
    \multirow{2}{*}{\textbf{Agg.}} & 
    \multirow{2}{*}{\textbf{Training Data}} &
    \multicolumn{2}{c}{Camelyon16} & \multicolumn{2}{c}{TCGA-NSCLC} & \multicolumn{2}{c}{TCGA-BRCA} \\
    \cmidrule(lr){3-4}  \cmidrule(lr){5-6} 
    \cmidrule(lr){7-8}  
     &   & Acc & AUC & Acc & AUC & Acc & AUC \\ 
    \midrule
    \multirow{3}{*}{\textbf{ABMIL}} & $10\%$ &  76.74 & 66.17 & 89.05 & 94.66 & 69.70 & 70.62 \\
     & $10\%$ merged & 79.84 & 70.55 & 90.48 & 95.47  & 68.48 & 82.21 \\
     & $50\%$ merged & 67.44 & 74.73  & 89.52 & 94.53 & 79.39 & 86.47 \\
    \midrule
    \multirow{3}{*}{\textbf{DSMIL}} & $10\%$ &  71.32 & 60.60 & 81.43 & 92.71 & 55.15 & 70.92 \\
     & $10\%$ merged & 76.74 & 63.88 & 81.90 & 93.76 & 89.79 & 83.26 \\
     & $50\%$ merged & 78.29 & 77.46 & 89.05 & 95.10 & 66.06 & 87.29 \\
    \midrule
    \multirow{3}{*}{\textbf{CLAM-SB}} & $10\%$  & 74.42 & 59.41 & 88.57 & 93.12 & 78.18 & 80.07 \\
     & $10\%$ merged  & 78.29 & 79.11 & 89.52 & 95.35 & 83.64 & 82.70 \\
     & $50\%$ merged & 82.17 & 81.61 & 92.86 & 96.11 & 73.94 & 84.15 \\
    \bottomrule
    \end{tabular}
    }
\caption{Results (\%) of Bag-level class classification on Camelyon16, TCGA-NSCLC and TCGA-BRCA with feature extractor pre-trained on the single and merged dataset. Acc and AUC are reported.}
\label{scaling_dataset}
\end{table}

\paragraph{Experiment settings}
To investigate the scaling laws related to data size, we conduct experiments using the SimMIL framework on the merged dataset. We merge the training sets of the Camelyon16, TCGA-NSCLC, and TCGA-BRCA datasets for pre-training, resulting in a six-class classification task. We then conduct downstream experiments on their respective test sets, comparing against the baselines trained on individual datasets.
Given computational constraints, we use only 10\% and 50\% of the training data from each dataset for the scaling experiments. For all settings, we pre-train for 100 epochs, with all other settings consistent with the benchmarking experiments.

\paragraph{Results}
~\cref{scaling_dataset} presents the results of feature extractors trained on single and merged datasets. With 10\% of the training data, the model pre-trained on the merged dataset outperforms those pre-trained on individual datasets across all settings. Consistent performance gains are also observed as the merged datasets scale from 10\% to 50\%.

This suggests that it is feasible to pre-train the feature extractor once using SimMIL for all downstream tasks, leading to a stronger feature extractor. Overall, the results demonstrate the scalability of SimMIL.

\subsection{Ablation Study}
\label{ablation}

\begin{table}[t!]
    \centering
{
    \begin{tabular}{ccccccccc}
        \toprule    
        \multirow{2}{*}{\textbf{Aug.}}  & \multirow{2}{*}{\textbf{MLP}} & \multirow{2}{*}{\textbf{Loss}} & \multicolumn{2}{c}{Camelyon16} & \multicolumn{2}{c}{TCGA-NSCLC}  & \multicolumn{2}{c}{TCGA-BRCA} \\
        \cmidrule(lr){4-5} \cmidrule(lr){6-7} \cmidrule(lr){8-9}
        & & & Acc & AUC & Acc & AUC & Acc & AUC \\
        \midrule
        & & & 73.13 & 59.65 & 85.72 & 92.24 & 66.06 & 58.58 \\
        \checkmark & & & 79.07 & 68.19 & 93.97 & 97.10 & 81.82 & 89.48\\
        \checkmark & \checkmark & & 78.81 & 78.23 & 93.65 & 97.19 & 88.48 & 92.08 \\
        \checkmark & \checkmark & \checkmark & 82.39 & 84.35 & 93.81 & 97.45 & 82.22 & 91.22\\
        \bottomrule
    \end{tabular}}
\caption{Ablation study of three modules. Acc and AUC are reported.}
\label{ablation_study}
\end{table}

\paragraph{Experiment settings}
We conduct ablation experiments to verify the effect of three modules: strong data augmentation, the MLP prediction head, and the symmetric loss function. CLAM-SB is used as the aggregation network because of its stability and strong performance.

\paragraph{Results}
~\cref{ablation_study} presents the results of the ablation study. We report the average results across three experiments in the table. We can see that 1) Strong augmentation brings significant improvement, with average gains of $8.54\%$, $4.86\%$, and $30.90\%$ in AUC across the three datasets. 2) Both the MLP prediction head and SCE loss bring benefits on Camelyon16, while they have a relatively smaller impact on TCGA-NSCLC and TCGA-BRCA. This is primarily due to the varying difficulty of the tasks: when the tasks are relatively easy, even the baseline model, SimpleMIL, can achieve promising results (e.g., on TCGA-NSCLC), which aligns with the observations in~\cite{dsmil}. Overall, the results show that SimMIL can serve as a simple and general framework for representation learning on MIL, especially for difficult tasks.


\subsection{Visualization on Real-World WSIs}

\begin{figure}[h]
    \centering
    \begin{subfigure}{0.3\linewidth}
      \centering
      \includegraphics[width=\linewidth]{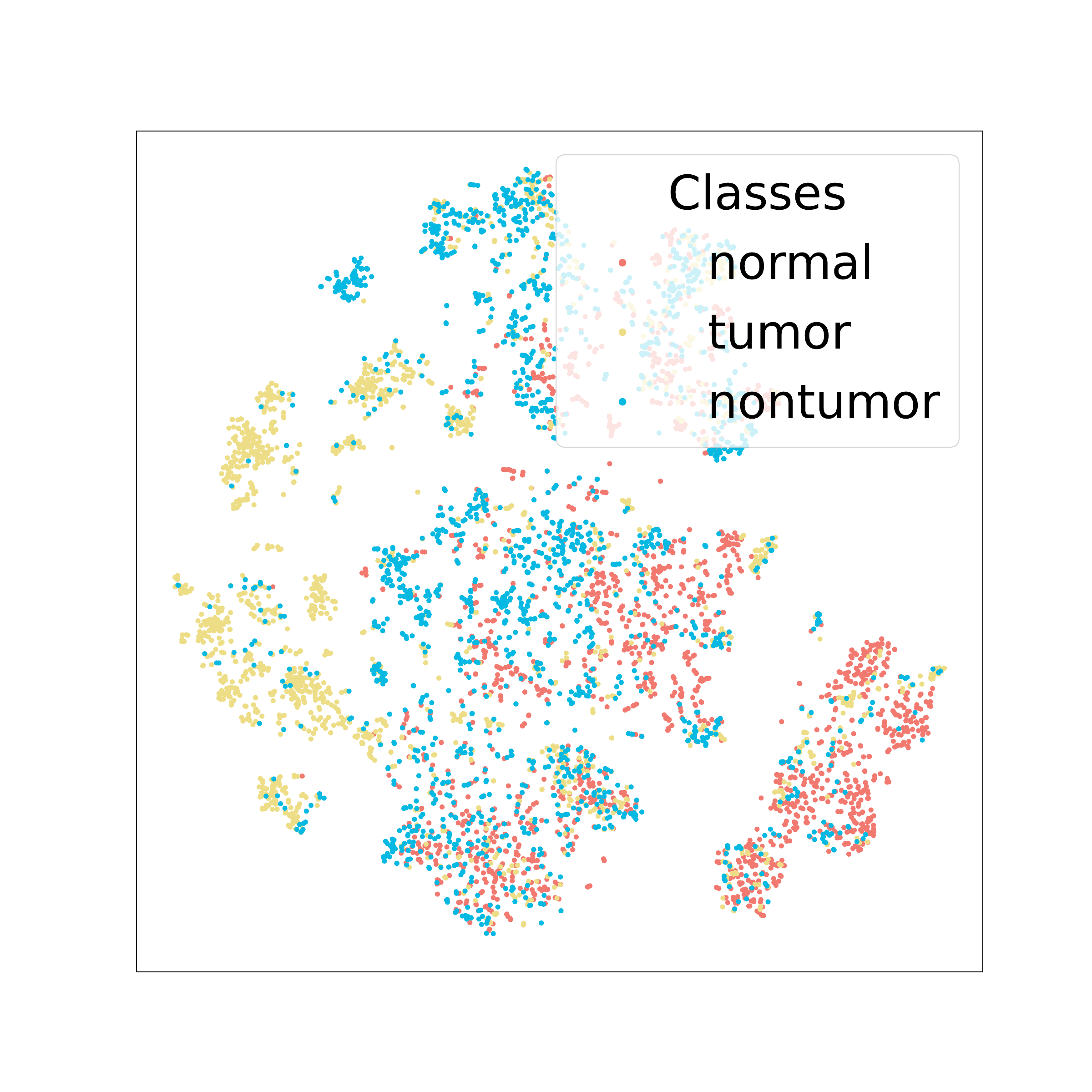}
      \subcaption{SimMIL}
      \label{tsne_wsl}
    \end{subfigure}
    \hfill
    \begin{subfigure}{0.3\linewidth}
      \centering
      \includegraphics[width=\linewidth]{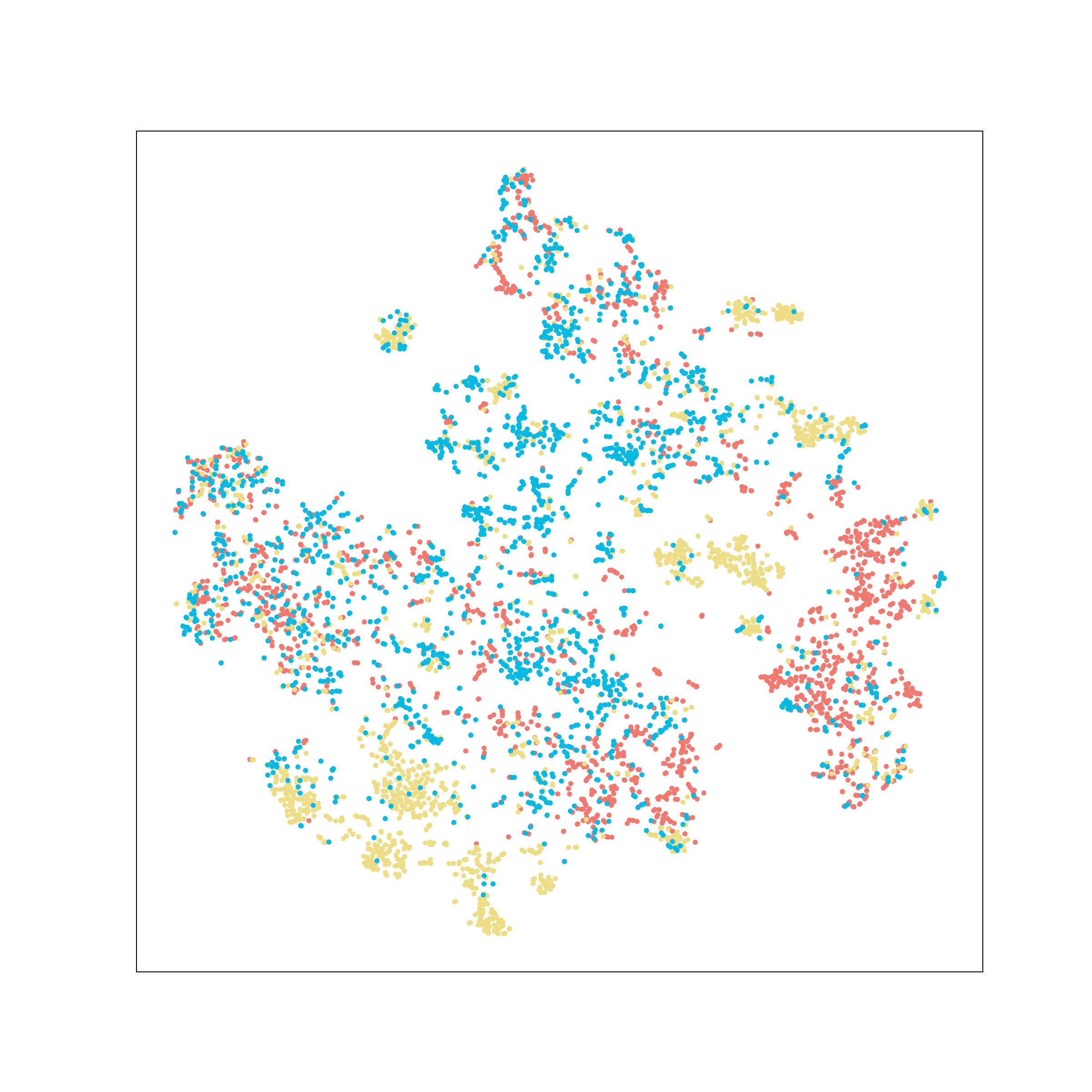}
      \subcaption{SimpleMIL}
      \label{tsne_simple}
    \end{subfigure}
    \hfill
    \begin{subfigure}{0.3\linewidth}
      \centering
      \includegraphics[width=\linewidth]{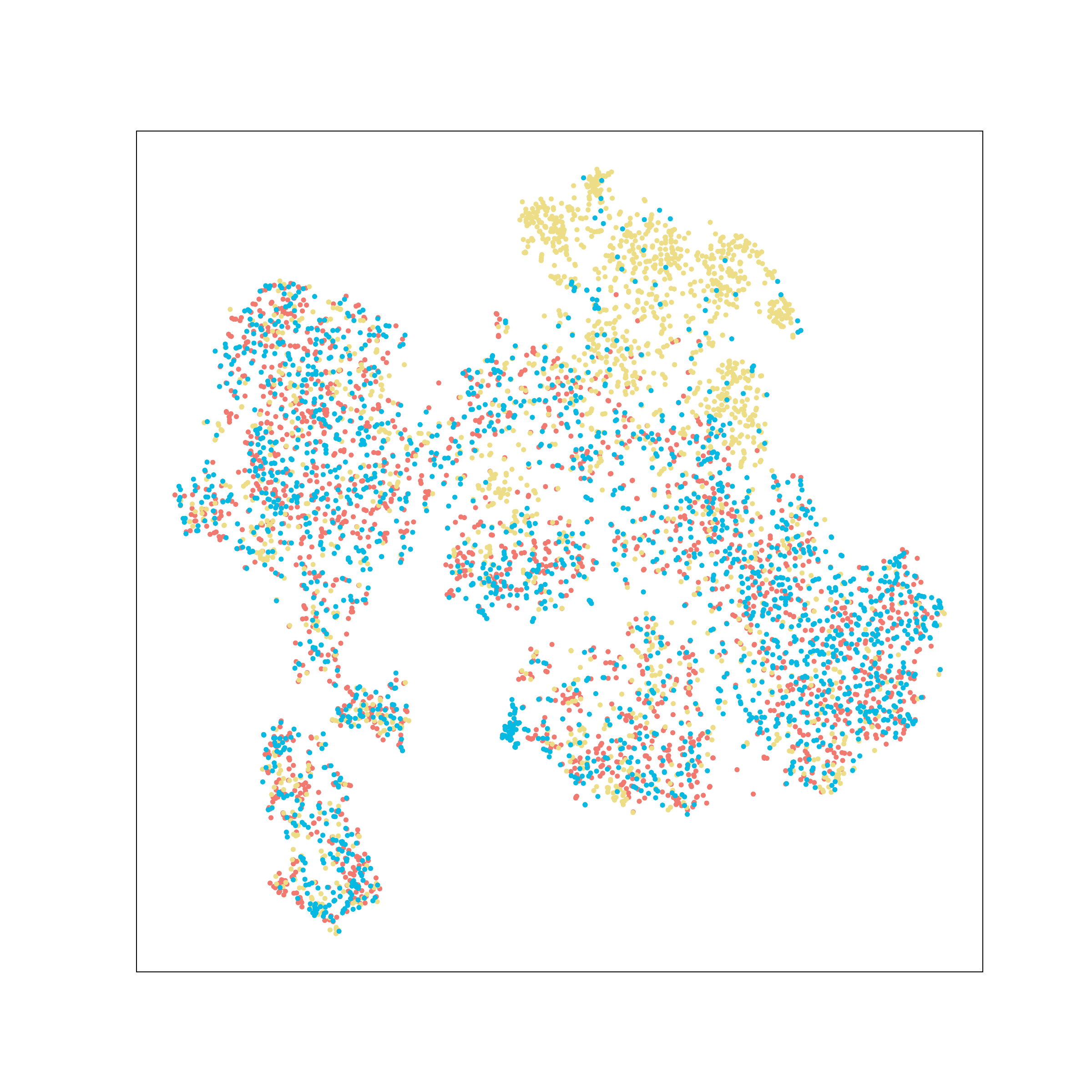}
      \subcaption{ImageNet}
      \label{tsne_ip}
    \end{subfigure}
  
    \begin{subfigure}{0.3\linewidth}
      \centering
      \includegraphics[width=\linewidth]{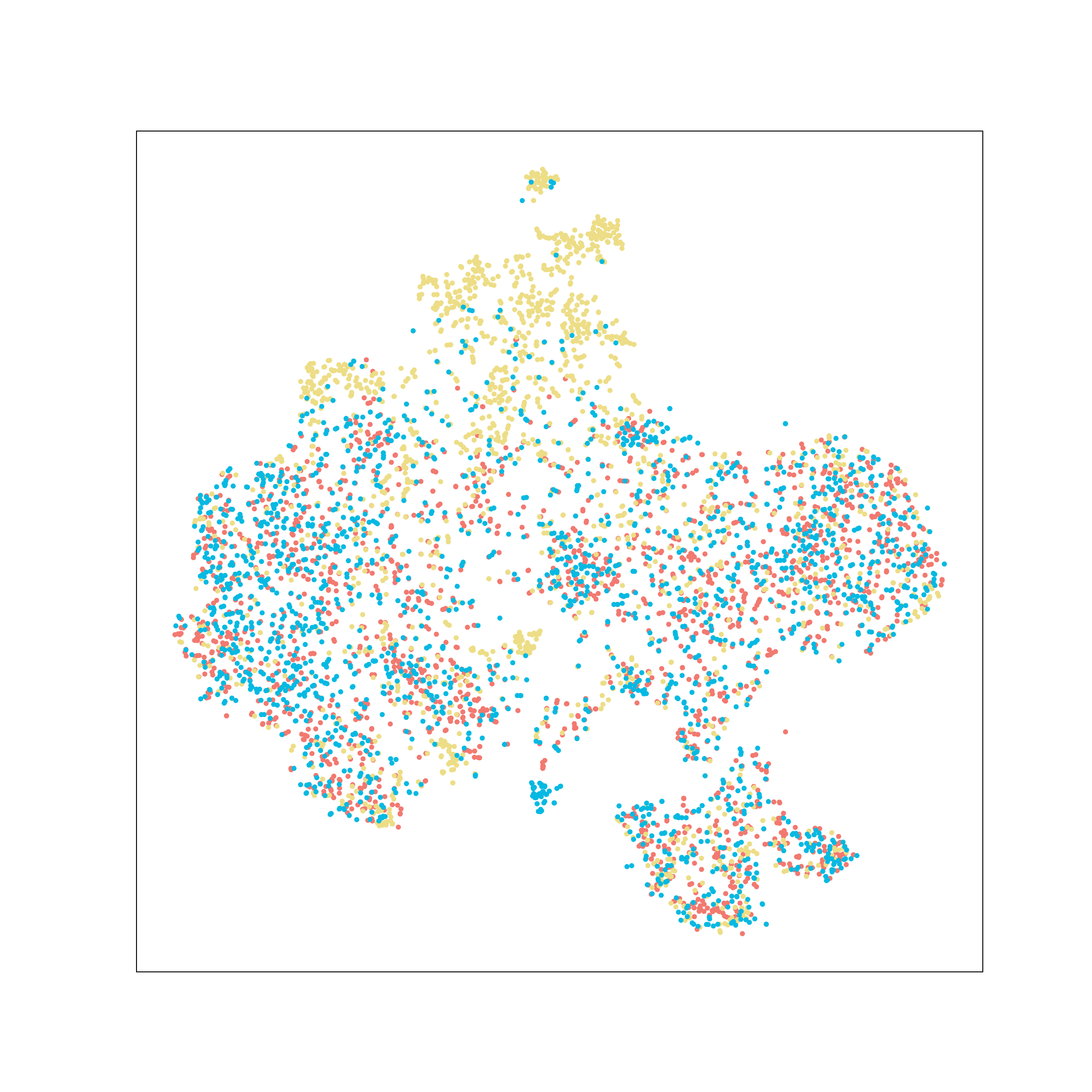}
      \subcaption{MoCo v2}
      \label{tsne_moco}
    \end{subfigure}
    \hfill
    \begin{subfigure}{0.3\linewidth}
      \centering
      \includegraphics[width=\linewidth]{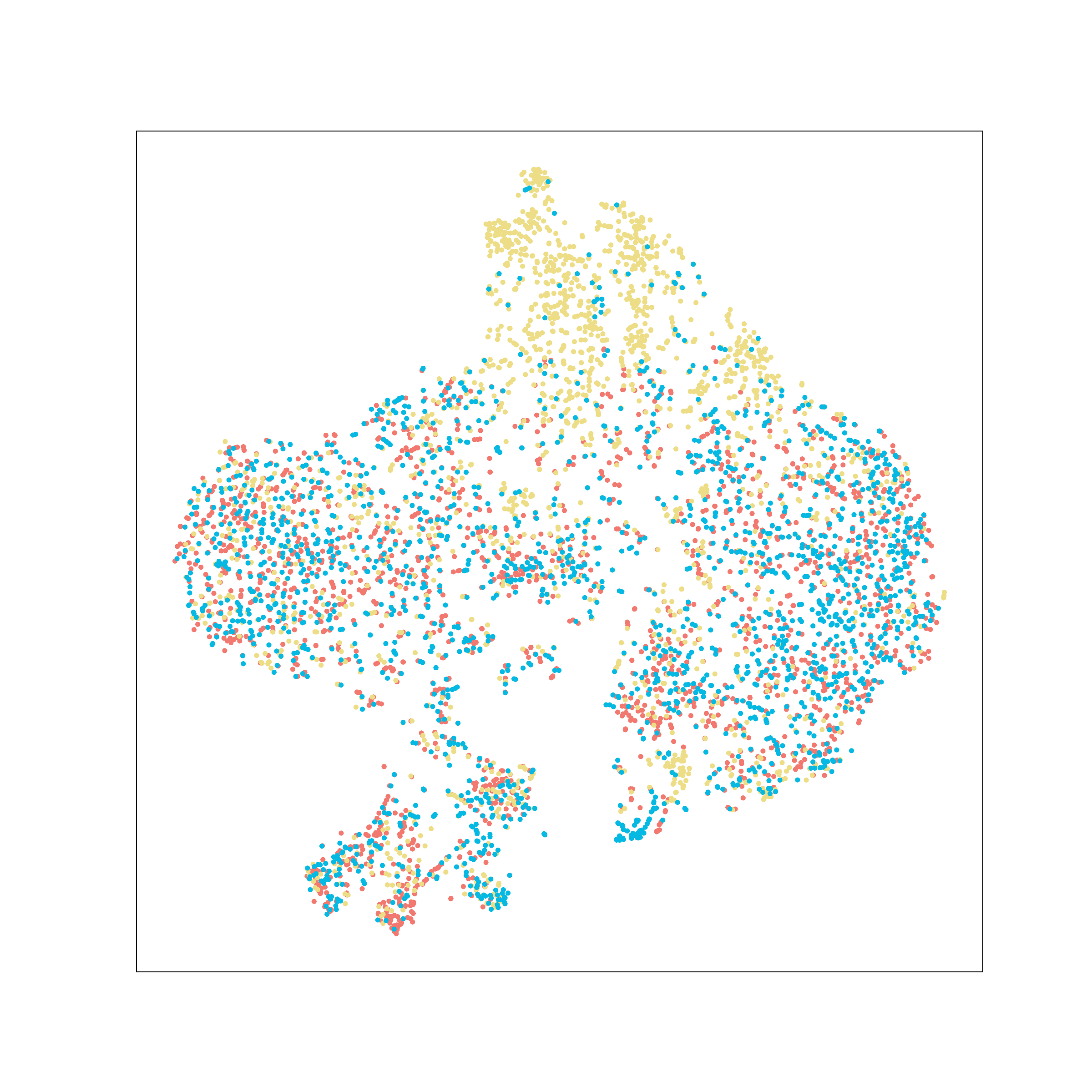}
      \subcaption{SimCLR}
      \label{tsne_simclr}
    \end{subfigure}
    \hfill
    \begin{subfigure}{0.3\linewidth}
      \centering
      \includegraphics[width=\linewidth]{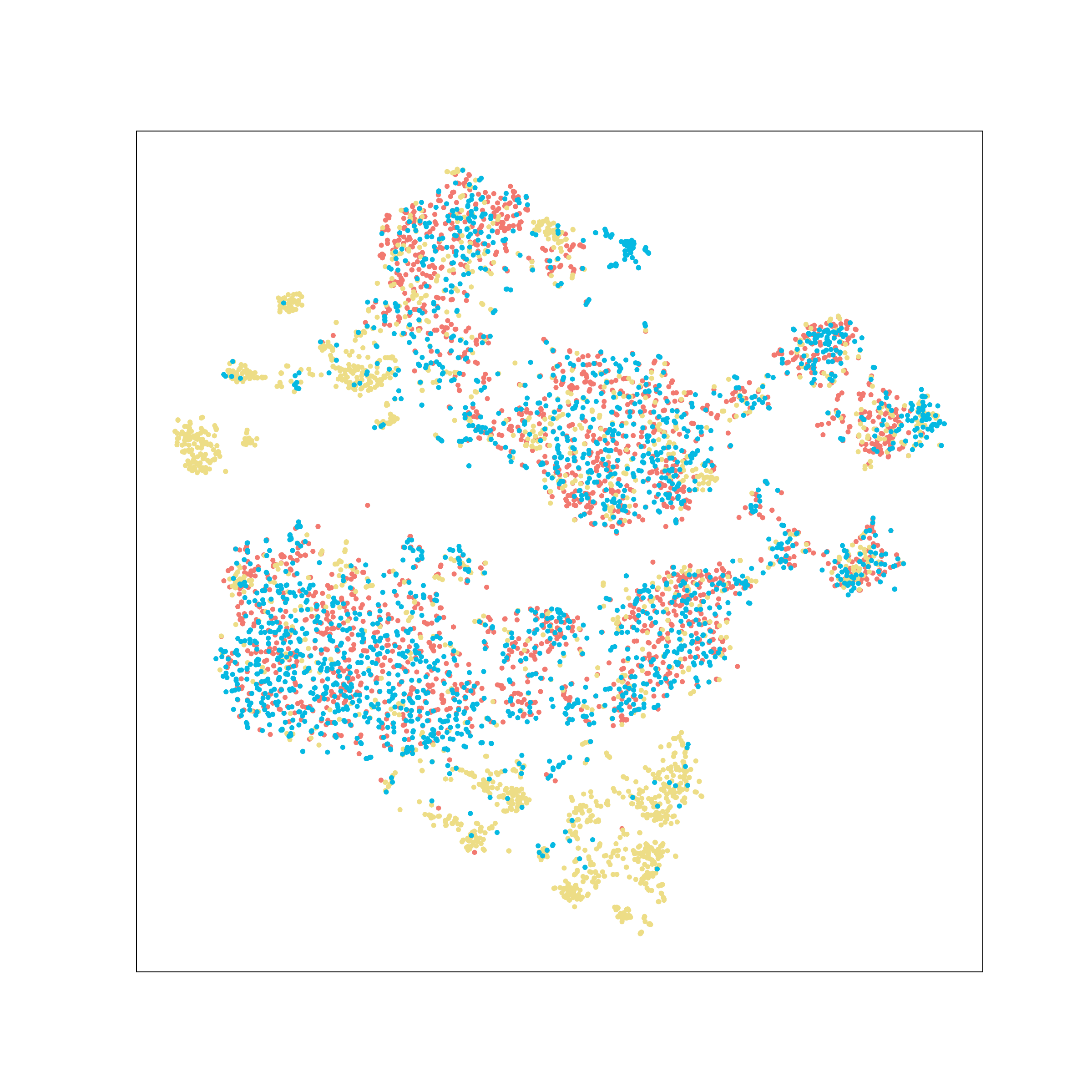}
      \subcaption{CTransPath}
      \label{tsne_ctrans}
    \end{subfigure}
    
  \caption{The t-SNE visualization of instance level features. Three classes, normal, tumor and nontumor, denote positive instances in positive bags, negative instances in negative and positive bags, respectively. Features are generated by different pre-training methods on Camelyon16. Legend is the same for 6 subfigures.
  }
  \label{fig:visual}
  \end{figure}

Figure~\ref{fig:visual} is the t-SNE visualization of instance-level features on Camelyon16, which is the only dataset with instance-level labels available.
Compared to SimpleMIL, the features extracted with SimMIL succeed in telling positive instances apart from negative ones. 
Compared with other baselines, SimMIL produces more separable feature representations among classes of tumor and others, indicating the superior effectiveness of SimMIL. 


\begin{figure}[ht!]
    \normalsize
      \centering
      \includegraphics[width=\textwidth]{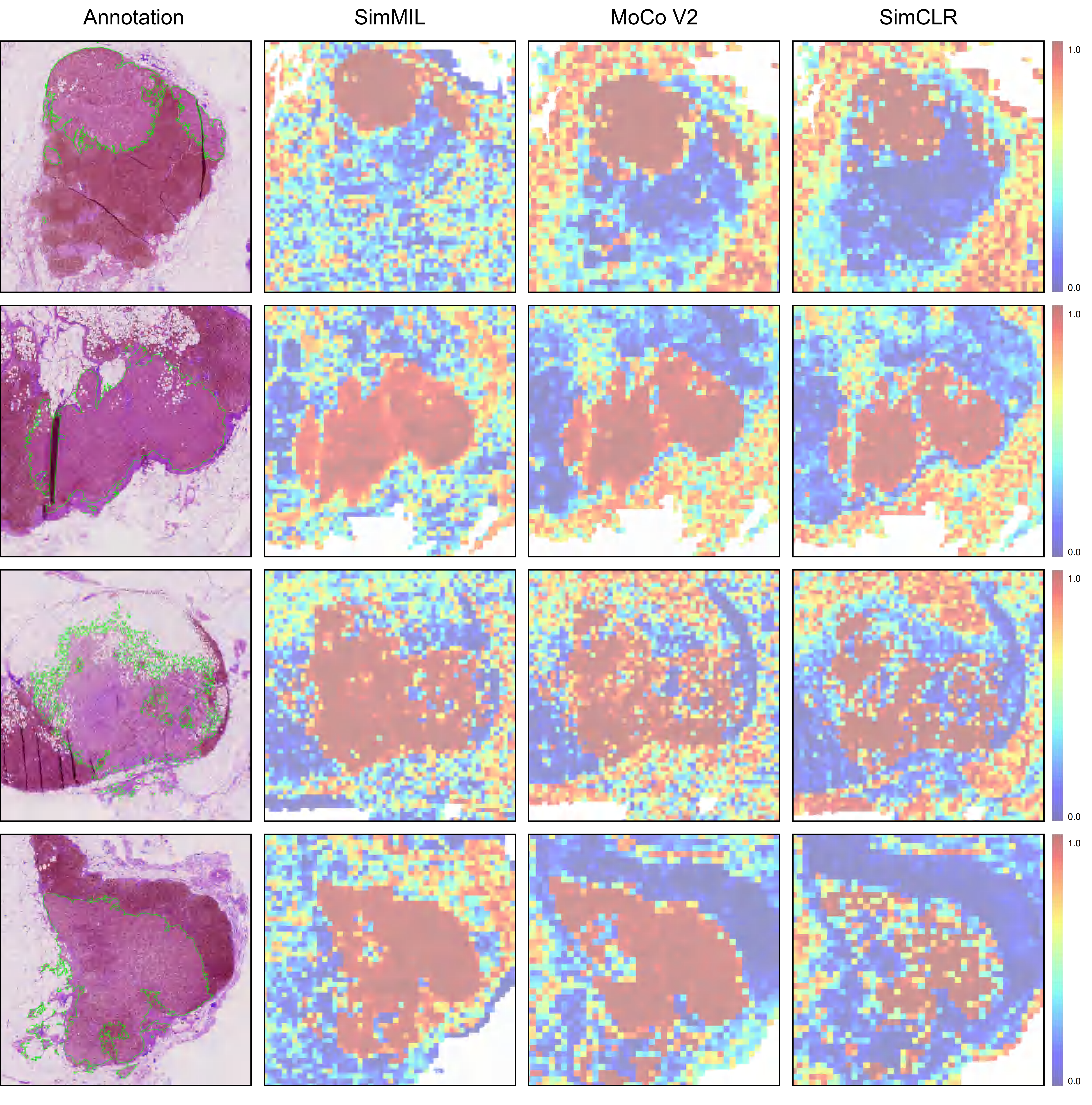}
      \caption{Attention map of CLAM-SB aggregator using features from different pre-training paradigms. The WSI is from Camelyon16.}
      \label{fig:attention1}
\end{figure}
    
\begin{figure}[ht!]
    \normalsize
    \centering
    \includegraphics[width=\textwidth]{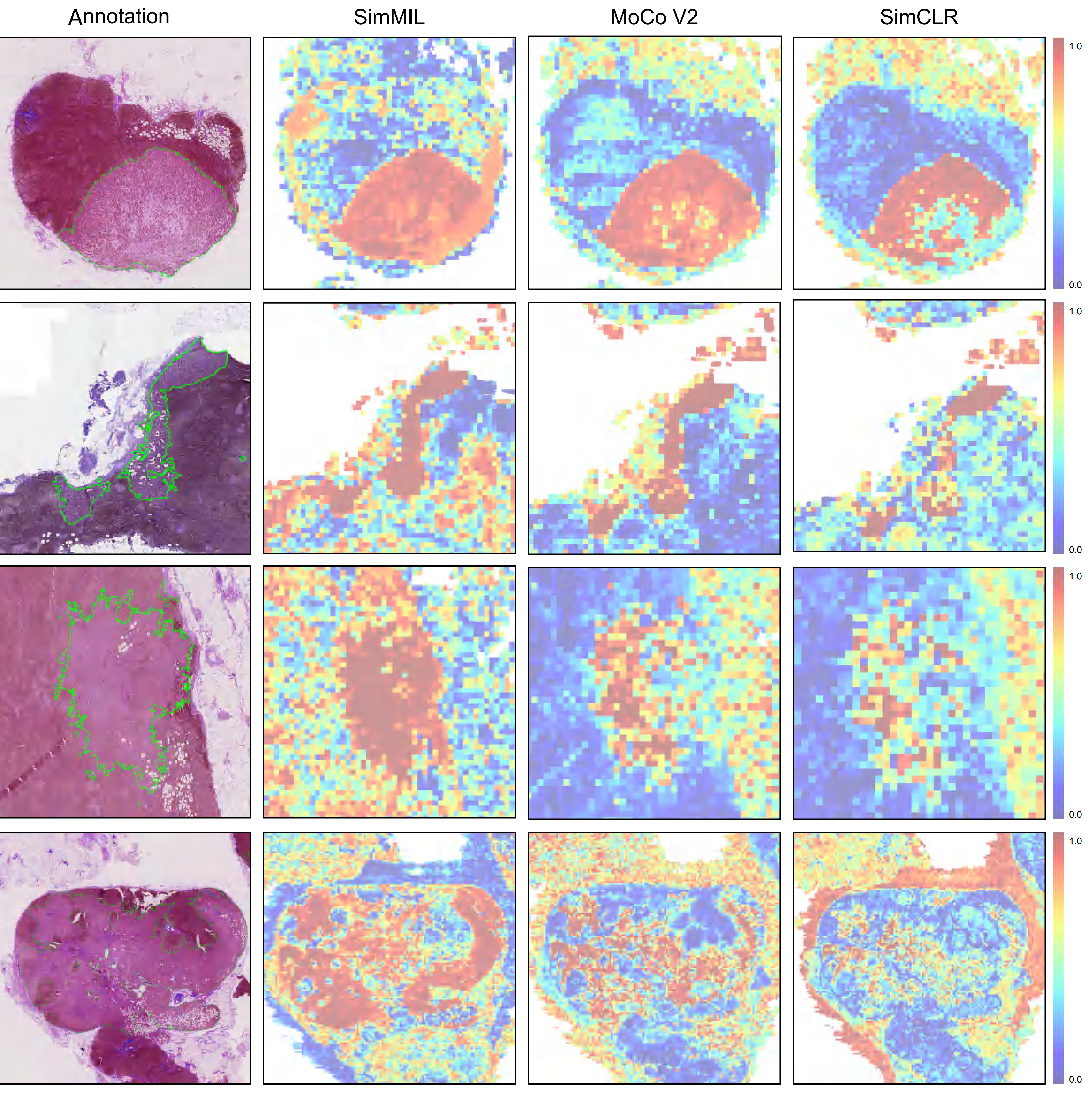}
    \caption{Attention map of CLAM-SB aggregator using features from different pre-training paradigms(Continued). The WSI is from Camelyon16.}
    \label{fig:attention2}
    \end{figure}

Following CLAM-SB~\cite{clam}, we demonstrate the attention maps using features from different pre-training paradigms in ~\cref{fig:attention1} and ~\cref{fig:attention2}. 
In the first sub-column, the area enclosed by the green line is the tumor region, while the other columns represent attention heatmaps generated by different pre-training methods.
The attention trained with our SimMIL is more focused on the tumor regions in the WSI because its focus is on the semantics of the patch rather than its 
graphical features.

\section{Conclusion}
In this work, we present a weakly supervised pre-training framework, SimMIL, for representation learning in multi-instance learning tasks. This framework enables training the feature extractor via bag-to-instance label propagation without extra requirements. This approach significantly narrows the gap between pre-training and downstream MIL tasks. We conducted extensive experiments with various MIL benchmarks on different tasks, and our results validate the effectiveness, compatibility, and scalability of SimMIL. A key insight is that instance-level MIL can function as a strong feature extractor for bag-level MIL methods, motivating a reconsideration of how to combine the two counterparts. In the future, we plan to design more pretext tasks to effectively inject task-specific knowledge, as well as explore additional MIL tasks with different assumptions. We hope that SimMIL can serve as a strong baseline for future studies.

\section*{Acknowledgments}
This work was supported in part by National Natural Science Foundation of China (No. 62171282) and Science and Technology Commission of Shanghai Municipality (No. 22DZ2229005).

\appendix
\section{More details of preliminary}
As briefly described in the main manuscript, we conduct preliminary experiments on the NCTCRC dataset~\cite{NCTCRC} and construct the NCTCRC-BAGS dataset by the instances in NCTCRC. 

\begin{table}[ht]
    \centering
    \begin{tabular}{ccc}
        \toprule    
        \multirow{2}{*}{Assumption}  & \multicolumn{2}{c}{Number of positive / negative bags} \\
        \cmidrule(lr){2-3}
        & Train & Test \\
        \midrule
        Standard & 1030 / 957 & 88 / 51 \\
        \bottomrule
    \end{tabular}
\caption{The summary of the created NCTCRC-BAGS dataset under MIL assumptions with number of positive and negative bags in training and test sets.}
\label{NCTCRC-Sumary}
\end{table}

The \textbf{NCTCRC} dataset consists of two subsets named \verb=NCR-CRC-HE-100K= and \verb=CRC-VAL-HE-7K= containing 100,000 and 7,180 non-overlapping image patches with $224\times224$ pixels at $0.5$ microns per pixel (MPP) from 9 tissue classes, which are adipose (\textit{ADI}), background (\textit{BACK}), debris (\textit{DEB}), lymphocytes (\textit{LYM}), mucus (\textit{MUC}), smooth muscle (\textit{MUS}), normal colon mucosa (\textit{NORM}), cancer-associated stroma (\textit{STR}), colorectal adenocarcinoma epithelium (\textit{TUM}). 

The \textbf{NCTCRC-BAGS} is a MIL dataset constructed following the MNIST-bags~\cite{abmil}. A bag in NCTCRC-BAGS is made up of 50 images taken without replacing, generating around 2,000 and 140 bags for training and test sets. For the downstream bag-level classification, we create the NCTCRC-BAGS under standard assumptions: A positive bag contains at least one instance from positive class. The summary of NCTCRC-BAGS under standard assumptions with the positive class \textit{TUM} is showed in ~\cref{NCTCRC-Sumary}.

\section{Scaling experiments on model size }

We conduct an additional ablation study to assess the performance of our SimMIL with different architectures. The SimMIL feature extractor was scaled from ResNet18 to ResNet50, using the same hyper-parameter settings. Subsequently, we further compare the downstream results with other models that utilized ResNet50 as the feature extractor.
We choose two stronger pre-trained models released by~\cite{kang2023benchmarking}, which are pre-trained by MoCo v2~\cite{MoCoV2} and SwAV~\cite{swav} on  $32.6M$ patches for 200 \textit{ImageNet epochs}~\cite{ImageNetEpoch}.

\begin{table}[ht!]
    \centering
    \resizebox{\linewidth}{!}
    {
    \begin{tabular}{ccccccccc}
    \toprule
    \multirow{2}{*}{\textbf{Agg.}} & 
    \multirow{2}{*}{\textbf{Method}} &
    \multirow{2}{*}{\textbf{Arch.}} &
    \multicolumn{2}{c}{Camelyon16} & \multicolumn{2}{c}{TCGA-NSCLC} & \multicolumn{2}{c}{TCGA-BRCA} \\
    \cmidrule(lr){4-5}  \cmidrule(lr){6-7} 
    \cmidrule(lr){8-9}  
     &  &  & Acc & AUC & Acc & AUC & Acc & AUC \\ 
     \midrule
    \multirow{4}{*}{\makecell{Max \\ Pooling}} & MoCo V2 & ResNet50 
    & 70.54 & \underline{74.13} & 89.52 & 95.28 & 66.67 & 75.26 \\
    & SwAV & ResNet50 & \underline{73.64} & 69.03 & \underline{89.52} & \underline{96.64} & 78.79 & 85.37 \\
    & SimMIL\footnotesize{(ours)} & ResNet18 & \textbf{80.62} & \textbf{79.63} & 89.52 & 95,91 & \textbf{87.88} & \underline{91.02} \\
    & SimMIL\footnotesize{(ours)} & ResNet50 & 72.87 & 65.23 & \textbf{90.48} & \textbf{97.45} & \underline{87.27} & \textbf{93.44} \\
    \midrule
    \multirow{4}{*}{\makecell{Mean \\ Pooling}} & MoCo V2 & ResNet50 & 65.89 & \underline{58.21} & 80.48 & 87.00 & 63.64 & 72.61 \\
    & SwAV & ResNet50 & 68.22 & 53.00 & \underline{87.14} & \underline{93.76} & 80.00 & 81.86 \\
    & SimMIL\footnotesize{(ours)} & ResNet18 & \textbf{75.19} & \textbf{59.48} & 86.67 & 92.86 & \underline{84.85} & \underline{87.59} \\
    & SimMIL\footnotesize{(ours)} & ResNet50 & \underline{70.54} & 57.28 & \textbf{89.52} & \textbf{95.43} & \textbf{90.30} & \textbf{90.35} \\
    \bottomrule
    \end{tabular}
    }
    \caption{Results (\%) of scaling experiments on model size, comparing ResNet18 and ResNet50. Acc and AUC are reported.}
    \label{ablation_backbone}
\end{table}

~\cref{ablation_backbone} illustrates the linear MIL probing for various architectures. 
We can observe that 1) SimMIL consistently outperforms other approaches. 2) With the scaling of model size, the downstream performance of SimMIL increases. 3) Under the SimMIL framework, models pre-trained by ResNet18 perform better than those pre-trained by ResNet50 on the Camelyon16. We believe it comes from the over-fitting problem because Camelyon16 contains the fewest WSIs. 
In summary, these results demonstrate the potential for SimMIL with stronger architectures. 

\section{Additional results}

We present the results under another two attention based aggregation network, TransMIL~\footnote{\url{https://github.com/szc19990412/TransMIL}}~\cite{transmil} and DTFD-MIL~\footnote{\url{https://github.com/hrzhang1123/DTFD-MIL}}~\cite{dtfdmil} in ~\cref{additional_results}. The overall results remain consistent with the results in the main manuscript. SimMIL achieves the best results on TCGA-NSCLC, and comparable or even better results on Camelyon16 and TCGA-BRCA, comparing with baselines. 

\begin{table}[ht!]
    \centering
    \resizebox{\linewidth}{!}
    {
    \begin{tabular}{ccccccccc}
    \toprule
    \multirow{2}{*}{\textbf{Agg.}} & 
    \multirow{2}{*}{\textbf{Method}} &
    \multirow{2}{*}{\textbf{Arch.}} &
    \multicolumn{2}{c}{Camelyon16} & \multicolumn{2}{c}{TCGA-NSCLC} & \multicolumn{2}{c}{TCGA-BRCA} \\
    \cmidrule(lr){4-5}  \cmidrule(lr){6-7} 
    \cmidrule(lr){8-9}  
     &  &  & Acc & AUC & Acc & AUC & Acc & AUC \\ 
     \midrule
    \multirow{6}{*}{\makecell{TransMIL}} & ImageNet & ResNet18 
    & 76.74 & 74.78 & 82.38 & 88.91 & \textbf{84.24} & 85.58 \\
    & MoCo V2 & ResNet18  & 80.62 & 75.13 & \underline{90.48} & 94.60 & 74.55 & 76.79 \\
    & SimCLR & ResNet18 & 75.19 & 71.29 & 88.57 & 91.68 & 76.36 & 79.89 \\
    & CLIP & ResNet50  & 75.19 & 69.11 & 84.29 & 91.38 & 82.42 & \textbf{90.40} \\
    & SRCL & CTransPath  & \textbf{86.05} & \textbf{87.35} & 89.05 & \underline{96.04} & 80.61 & \underline{89.60} \\
    & SimMIL\footnotesize{(ours)} & ResNet18  & \underline{81.40} & \underline{79.95} & \textbf{94.76} & \textbf{97.18} & \underline{82.42} & 83.37 \\
    \midrule
    \multirow{6}{*}{\makecell{DTFD-MIL}} & ImageNet & ResNet18 & 84.50 & 83.90 & 88.10 & 92.10 & 71.52 & 80.59 \\
    & MoCo V2 & ResNet18 & 54.26 & 55.99 & 77.62 & 86.24 & 74.55 & 74.86 \\
    & SimCLR & ResNet18 & 78.29 & 81.56 & 87.62 & 93.44 & 82.42 & 85.70 \\
    & CLIP & ResNet50  & 67.44 & 56.32 & 71.90 & 76.96 & 58.18 & 67.17 \\
    & SRCL & CTransPath  & \textbf{93.80} & \textbf{95.58} & \underline{90.00} & \underline{96.46} & \underline{89.70} & \textbf{93.72} \\
    & SimMIL\footnotesize{(ours)} & ResNet18 & \underline{93.02} & \underline{91.10} & \textbf{94.29} & \textbf{97.58} & \textbf{90.30} & \underline{92.48} \\
    \bottomrule
    \end{tabular}
    }
    \caption{Results (\%) of bag classification using TransMIL and DTFD-MIL as aggretators. Acc and AUC are reported.}
    \label{additional_results}
\end{table}

\section{More implementation details}

Taking into account the reproducibility of our experiment, we offer the implementation details for dataset preparation, augmentation, hyperparameters in pre-training, and downstream experiments.

\subsection{Preparation of datasets}

For Camelyon16 and TCGA-NSCLC, we directly utilize the patches released by other works~\cite{dsmil, IMIL}, while for TCGA-BRCA, we follow CLAM~\cite{clam} to segment the foreground regions of WSIs and exclude the WSIs not belonging to the two subtypes. 
For fine-tuning the second stage of HIPT, we follow the CLAM to segment $4096\times4096$ regions from TCGA-NSCLC and TCGA-BRCA. Then we divide each region to $256\times256$ patches and use the released model of stage one to get a 2D feature grid of $16\times16\times384$. 

\subsection{Pre-training details}

\paragraph{Module details}

\subparagraph{Augmentation} 
For all the images in our experiments, we apply the augmentation scheme following MoCo V2~\cite{MoCoV2}:

\begin{itemize}
    \item \textbf{Random resize crop}: images are cropped and resized to $224\times224$.
    \item \textbf{Weak color jittering ($p=0.8$)}: the brightness, contrast, saturation, and hue of images are randomly adjusted with a strength of $0.4, 0.4, 0.4, 0.1$, respectively.
    \item \textbf{Color dropping $(p=0.2)$}: the color of images are converted randomly to grayscale.
    \item \textbf{Random Gaussian Blur $(p=0.5)$}: images are randomly applied Gaussian filter.
    \item \textbf{Random horizontal flip $(p=0.5)$}: images are randomly applied horizontal flip.
\end{itemize}

\subparagraph{MLP prediction head}

The implementation of the MLP prediction head after the backbone (ResNet18 or ResNet50) follows BYOL~\cite{BYOL}, containing two linear layer, one batch normalization layer and one activation layer using ReLU. The dimension of hidden layer is set to be 128 for ResNet18 and 512 for ResNet50.

\paragraph{Pre-training hyper-parameters}

The pre-training phase of SimMIL implementation is based on the SimpleMIL~\cite{SimpleMIL}, which directly propagates bag-level labels to instances. We introduce three modules into the SimpleMIL framework, as described in the main manuscript. For benchmark experiments on the three datasets, we utilize a SGD optimizer with no weight decay, a momentum of 0.9 and a batch size of 256 on 4 GPUs. We set the initial learning rate to $1\times10^{-3}$ and train with a stepwise learning scheduler for 100 epochs on TCGA-NSCLC and TCGA-BRCA, 200 epochs on Camelyon16. The learning rate decreases at the 60 and 80 epochs when we train 100 epochs and at 120 and 160 epochs for 200 epochs totally. For fine-tuning CTransPath and HIPT, we set the initial learning rate to $5\times10^{-4}$ and train with a cosine annealing scheduler for 1 epoch, 5 epochs and 10 epochs.

\subsection{Downstream details}

\begin{table}[ht!]
    \centering
    \begin{tabular}{cccc}
        \toprule    
        Aggregator  & Scheduler & LR & weight decay \\
        \midrule
        Max Pooling & Adam & $1\times10^{-4}$ & $1\times10^{-4}$ \\
        Mean Pooling & Adam & $1\times10^{-4}$ & $1\times10^{-4}$ \\
        ABMIL & Adam & $1\times10^{-4}$ & $1\times10^{-4}$ \\
        DSMIL & Adam & $1\times10^{-4}$ & $1\times10^{-4}$ \\
        CLAM-SB & Adam & $1\times10^{-4}$ & $1\times10^{-5}$ \\
        \bottomrule
    \end{tabular}
\caption{The summary of the hyper-parameters using in the training of different aggreation networks.}
\label{aggregation_param}
\end{table}

\paragraph{Aggregator details}

For the downstream task we use two non-parametric aggregators and five attention based aggregators to validate the performance of feature extractor. 

\begin{itemize}
    \item \textbf{Non-parametric aggregator}: model contains only one linear layer after the non-learnable layer. We call it linear MIL probing.
    \item \textbf{ABMIL}: model uses a MLP to calculate the attention value of the instances in a bag. The code is available at \url{https://github.com/AMLab-Amsterdam/AttentionDeepMIL}. We use two linear layers and a activation layer to form the attention module.
    \item \textbf{DSMIL}: model uses the instance-level branch to score and selects the top-1 instance, then calculates the distances between this instance and others, using as the attention values. The code is available at \url{https://github.com/binli123/dsmil-wsi/tree/master}.
    \item \textbf{CLAM-SB}: model uses multi parallel attention branches to calculate bag-level representation for multi-classes, which trained with different set of high-attended regions. The code is available at \url{https://github.com/mahmoodlab/CLAM/tree/master}.
\end{itemize}

\paragraph{Downstream hyper-parameters}

The downstream part of SimMIL implementation is based on the code base of DSMIL~\cite{dsmil}, in which we first compute the instance-level features and then train an aggregation network. For all aggregators, we train for 50 epochs with an Adam optimizer. Other hyper-parameters are showed in ~\cref{aggregation_param}.




\bibliographystyle{elsarticle-num}
\bibliography{elsarticle-template-num.bbl}



\end{document}